# Hybrid consistency and plausibility verification of product data according to the EU regulation on food information to consumers (FIC)

Christian Schorr

Trier University of Applied Sciences, Environmental Campus Birkenfeld,
55761 Birkenfeld, Germany
c.schorr@umwelt-campus.de

**Abstract.** The labelling of food products in the EU is regulated by the Food Information of Customers (FIC). Companies are required to provide the corresponding information regarding nutrients and allergens among others. With the rise of e-commerce more and more food products are sold online. There are often errors in the online product descriptions regarding the FIC-relevant information due to low data quality in the vendors' product data base. In this paper we propose a hybrid approach of both rule-based and machine learning to verify nutrient declaration and allergen labelling according to FIC requirements. Special focus is given to the problem of false negatives in allergen prediction since this poses a significant health risk to customers. Results show that a neural net trained on a subset of the ingredients of a product is capable of predicting the allergens contained with a high reliability.

## 1 Introduction

### 1.1 Motivation

Since 25 October 2011, the Food Information to Customers (FIC) Regulation governs the labelling of foodstuffs in the European Union as Regulation No. 1169/2011 [EU 2011]. The FIC applies to food companies at every level of the food chain, insofar as their activities concern the provision of food information to consumers. In particular it applies to all foods intended for the final consumer, including foods supplied by mass caterers and foods intended for supply to mass caterers. The aim of the FIC is the provision of food information serves the overall protection of consumer health and interests by providing a basis for informed choice and safe use of food by final consumers, taking particular account of health, economic, environmental, social and ethical considerations.

In the context of this report, the following labels are relevant, which prepackaged food must bear in accordance with FIC.

- Nutrition declaration / nutrition labelling
- Ingredients and excipients (and their derivatives) listed in Annex II which cause allergies and intolerances
- List of ingredients, if more than one is contained

With the increasing importance of online trade in the food sector, it is important that the labelling required by the FIC is correctly indicated. In contrast to stationary trade, the customer in an e-shop cannot pick up the product and read the information on nutrients, allergens and ingredients provided by the manufacturer. Instead, they must be able to rely on the e-shop operator to correctly maintain the relevant information in his database and to reproduce it completely on his website. As product data is often of poor quality and subject to errors, there is a need for automated FIC compliance checks in the retail sector due to the huge amount of different products. We propose a hybrid approach of rule-based methods and machine learning algorithms to achieve this goal. The present report deals with two main topics within this context: rule-based consistency testing of nutritional values and ML-based prediction of contained allergens from the list of ingredients. In combination these two aspects cover the most important part of the labelling required by the FIC. The first chapter of this report introduces the problem and summarises the current state of science and technology. The second chapter explains the regulation on Food Information to Consumer (FIC) with a focus on nutrients and allergens. Questions to be considered regarding consistency and plausibility are discussed. The third chapter provides an overview of the product data used, explains necessary pre-processing steps and analyses the product data statistically. The fourth chapter introduces the problem of multi-label classification and presents various solutions and classification methods. The fifth chapter presents the results of the experiments and determines the best model based on these results. The sixth chapter summarises and explains the results. An outlook on further research is given.

## 1.2 State of the art

At the time of writing this report, there are no scientific publications on the determination of FIC compliance using machine learning methods. Rule-based checks for nutrient declaration compliance are straight-forward and used by most companies. The general problem of multi-label classification as it occurs in the prediction of allergens is the subject of ongoing research, though. A general overview can be found in [Zhang 2014, Zhang 2010, Tsoumakas 2007, Boutell 2004]. The approach of binary relevance has been studied in [Zhang 2018]. The topic of the Classifier Chains is dealt with in [Dembczyński 2012, da Silva 2014, Liu 2018, Tenenboim-Chekina 2013, Liu 2019].

# 2 EU Regulation Food Information to Customers (FIC)

## 2.1 Nutrition Declaration

According to FIC, every product has to display a declaration stating the amount the contained nutrients.

| Basis nutrient (abbreviation) | Energy Value [kJ/g] | Energy Value [kcal/g] | Implicit conversion factor |
|---|---|---|---|
| Carbohydrate (CH) | 17 | 4 | 4.25 |
| Polyols (POL) | 10 | 2.4 | 4.17 |
| Protein (PRO) | 17 | 4 | 4.25 |
| Fat (FAT) | 37 | 9 | 4.11 |
| Ethanol (ALC) | 29 | 7 | 4.14 |
| Organic acid (SFA/UFA) | 13 | 3 | 4.33 |
| Fibre (FIB) | 8 | 2 | 4.00 |
| Salt (SAL) | 0 | 0 | 1.00 |

*Table 1: Energy values of basic nutrients according to FIC (EU) Nr. 1169/2011 – Appendix XIV*

The FIC defines exactly the energy value of every nutrient. This value is specified both in kJ and in kcal. The resulting implicit conversion factor from kcal to kJ is variable for rounding reasons. Vitamins and minerals have no energy value. An overview can be found in Table 1. In addition, FIC clearly regulates the recommended daily reference intake of an average adult, defined per nutrient and with the accompanying unit of measurement (UoM). Table 2 lists the corresponding values:

| Nutrient / Vitamin / Mineral | Unit of measurement | Daily reference intakes [UoM] |
|---|---|---|
| Energy Value | kJ | 8400 |
| Energy Value | kcal | 2000 |
| Fat (FAT) | g | 70 |
| Saturates (SFA) | g | 20 |
| Carbohydrate (CH) | g | 260 |
| Sugar (SUG) | g | 90 |
| Protein (PRO) | g | 50 |
| Salt (SAL) | g | 6 |

*Table 2: Units of measurement and recommended daily reference intakes according to FIC (EU) Nr. 1169/2011 – Appendix XIV*

## 2.2 Allergen Labelling

FIC requires a mandatory labelling of the following allergens (Appendix II (FIC) Regulation (EU) Nr. 1169/2011):

- Cereals containing gluten
- Crustaceans
- Eggs
- Fish
- Peanuts
- Soybeans
- Milk and lactose
- Nuts
- Celery
- Mustard
- Sesame
- Sulphur dioxide and sulphites
- Lupine
- Molluscs

All products containing one or more of the allergens requiring labelling must have them marked in their list of ingredients. Either by indicating the ingredients containing allergens in capital letters or in bold type. In the product data from the SAP system there is no possibility to save individual ingredients in bold letters. Capital letters, however, are possible but are not stringently used depending on the company and master data manager. For the present report, therefore, the general case is assumed in which the allergenic ingredients are not specifically highlighted in the product data and therefore cannot be formally distinguished from allergen-free ingredients.

# 3 Data Model

## 3.1 Design of a FIC specific data model

The internal data structures of many companies do not follow a uniform standard. Product data is usually stored in an Enterprise Resource Planning system (ERP) such as SAP Retail. However, the naming individual attributes varies greatly. For this reason, a data model (see Fig. 1) has been developed specifically to meet the requirements of an FIC consistency check, on which all implemented algorithms in this report are based.

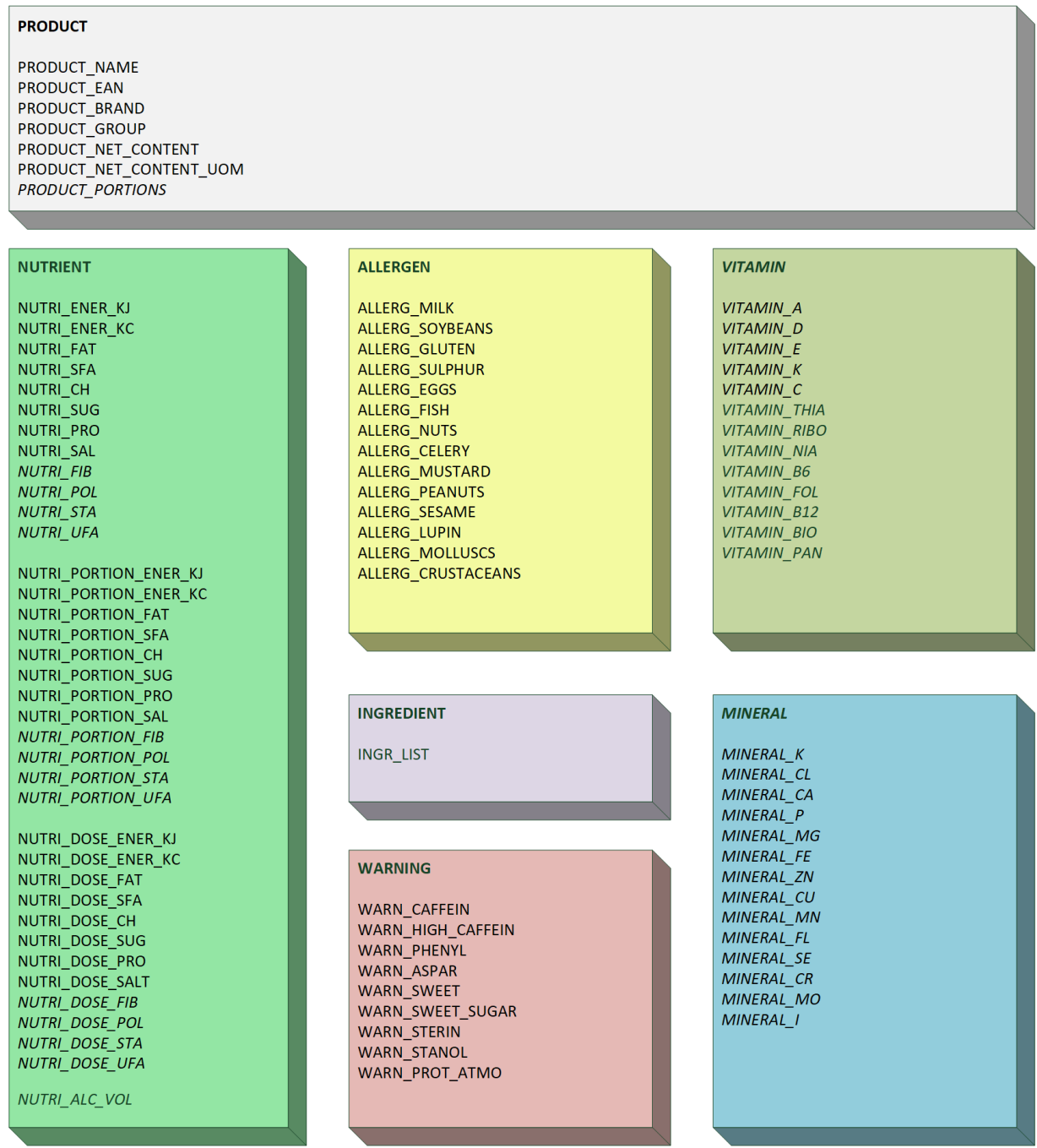

*Fig. 1: Data model FIC*

Instead of having to adapt the software individually for each company, using the FIC data model it is only necessary to assign the company-specific named attributes to those contained in the data model. This process can be carried out in a simple pre-processing step by means of a mapping table and thus leaves the underlying software untouched. The data model comprises seven areas. For the present technical report, the focus lies on the areas NUTRIENT, ALLERGEN and INGREDIENT. The number of items in the test data containing information on VITAMIN, MINERAL and WARNING is too small to justify inclusion. The data model nevertheless contains these areas, as it is intended to serve as a fundamental basis for all FIC-relevant checks. In addition to the verification methods developed here, this ensures an easy expandability by future checks.

### 3.1.1 Data Area Product

The data area Product comprises the general attributes of a product such as name, GTIN (Global Trade Item Number), brand, product group, net content, unit of measure of the net quantity and portions contained.

### 3.1.2 Data Area Nutrient

The nutrients of a product are summarised in the data area Nutrient. In addition to the absolute amounts, the data also includes the portion-wise information as well as the proportion of the recommended daily reference dose. As a further FIC relevant attribute, alcohol content per volume is also considered.

### 3.1.3 Data Area Allergen

The 14 allergens subject to labelling according to FIC form the allergen data area.

### 3.1.4 Data Area Vitamin

The data area Vitamin consists of the 13 vitamins relevant to FIC.

### 3.1.5 Data Area Mineral

The data area Mineral encompasses the 14 elements covered by FIC.

### 3.1.6 Data Area Ingredient

The list of ingredients of a product constitutes the data area Ingredient.

### 3.1.7 Data Area Warning

The data area Warning contains all warnings required by FIC, e.g. caffeine or artificial sweeteners.

## 3.2 Data Exploration

Our data set was provided by an European retailer consisting of 69076 products from the food sector. Among them 40063 products contained all FIC relevant attributes - nutrients, allergens and list of ingredients. In order to evaluate the data set for machine learning purposes, data exploration was conducted regarding the distribution of the difference nutrients and allergens.

### 3.2.1 Data Pre-processing

To obtain a processable data format, FIC-relevant information from different tables must be combined. The target format is the data model designed in 3.1. The individual aggregation steps depend on the company-specific data format and are not described in detail here for reasons of confidentiality.

### 3.2.2 Statistical Nutrient Analysis

Table 3 shows the number of products with a corresponding nutritional value greater than 0. A critical point is the unequal occurrence of the energy value information in kJ and kcal. According to the FIC, both values must be entered in pairs. However, only about half of all products in the data set have an energy value in kJ. Multi-chain alcohols and unsaturated fatty acids are rarely declared. Products containing starch are completely missing.

| Value | Number of products |
|---|---|
| ENER_KJ | 22912 |
| ENER_KC | 40897 |
| Fat (FAT) | 37505 |
| Saturated fatty acids (SFA) | 34197 |
| Carbohydrates (CH) | 37005 |
| Sugar (SUG) | 34773 |
| Protein (PRO) | 37141 |
| Salt (SAL) | 35274 |
| Fibres (FIB) | 9291 |
| Polyols (POL) | 77 |
| Starch (STA) | 0 |
| Unsaturated fatty acids (UFA) | 409 |

*Table 3: Nutrient distribution in the test data set*

### 3.2.3 Allergen Distribution

As expected, the allergens contained in the data set are not evenly distributed. Most of them can be assigned to the classes milk, gluten, soya and eggs. Exotic allergens such as lupine, molluscs and crustaceans, on the other hand, are only rarely found. The data set is therefore unbalanced which has to be taken into account during model training. Table 4 lists the number of products containing a specific allergen:

| Allergen | Number of products |
|---|---|
| Milk | 17283 |
| Soybeans | 10068 |
| Gluten | 12823 |
| Sulphur | 2870 |
| Eggs | 9897 |
| Fish | 2770 |
| Nuts | 7599 |
| Celery | 4609 |
| Mustard | 5275 |
| Peanuts | 2714 |
| Sesame | 2832 |
| Lupine | 802 |
| Molluscs | 606 |
| Crustaceans | 939 |

*Table 4: Allergen distribution in the test data set*

In order to analyse possible dependencies between allergens, heat maps were computed showing the relative and absolute frequencies of pairwise allergen occurrence (Fig. 2). As can be seen, the allergen milk, for example, occurs very frequently in combination with the allergen fish.

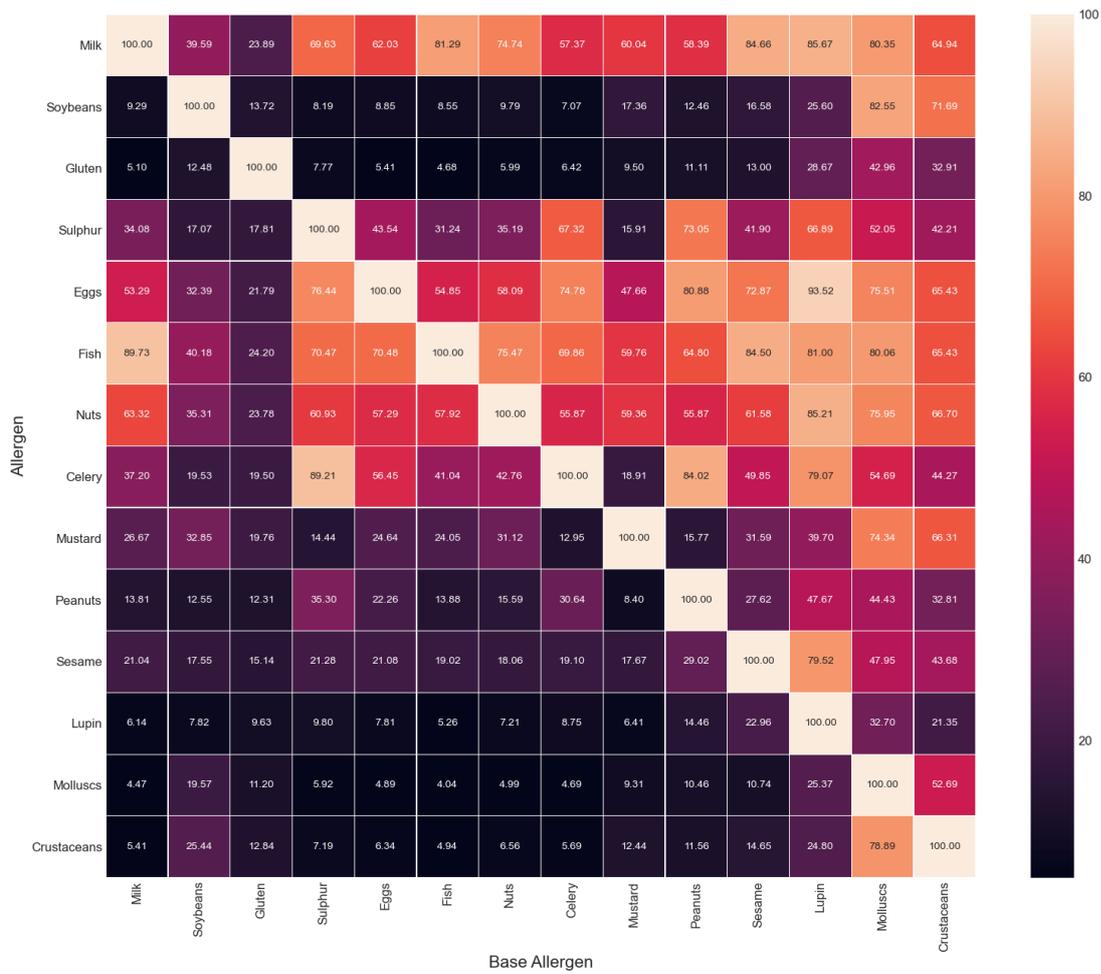

*Fig. 2: Relative frequency of pairwise allergen occurrence [%]*

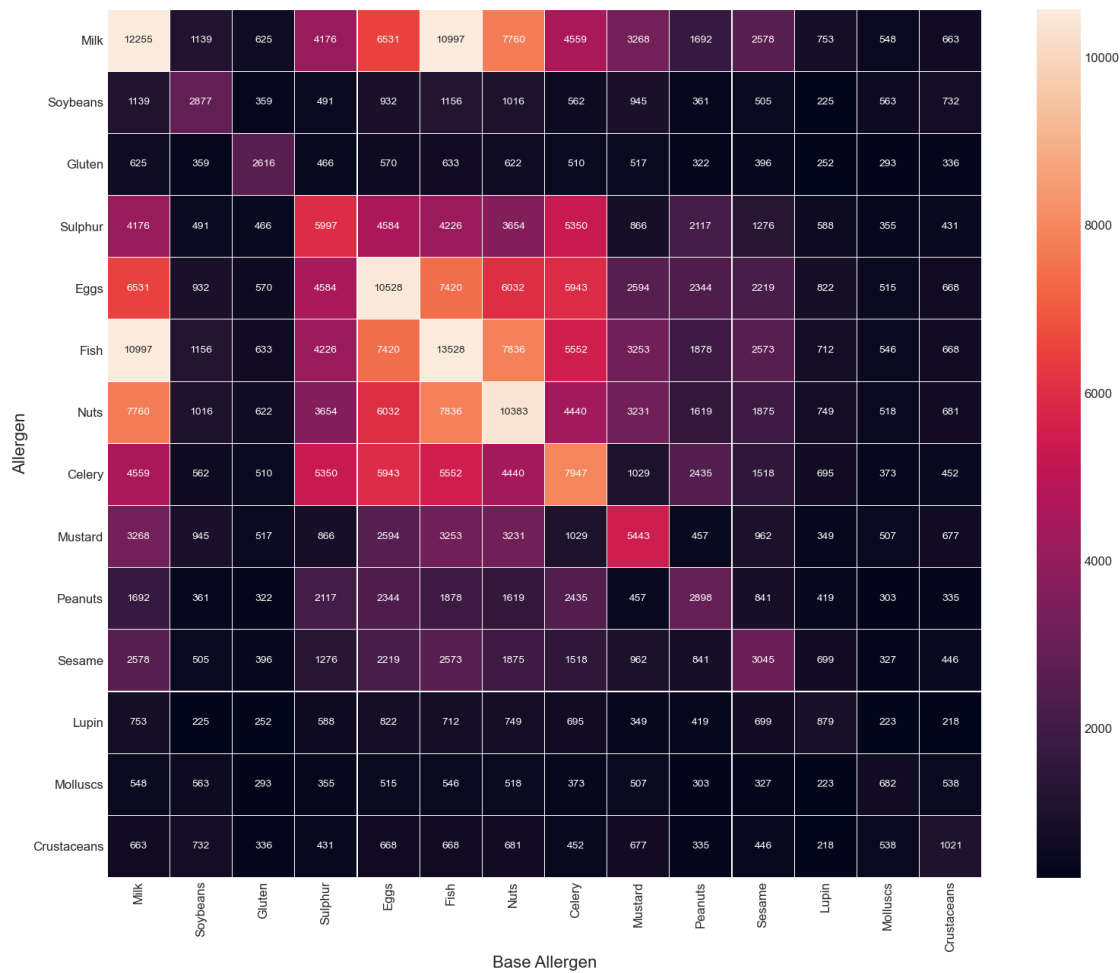

*Fig. 3: Absolute frequency of pairwise allergen occurrence*

# 4 Metrics and Classification Algorithms

## 4.1 Text transformations and classification algorithms

In order to render textual information usable for machine learning, the two text transformations Bag-of-words (BOW) and Term Frequency - Inverse Document Frequency (TF-IDF) [Jones 1972] were applied to the data set. For classification purposes we used neural networks (NN) [Schmidthuber 2015], support vector machines (SVM) [Cortes 1995], and the Random Forest (RF) algorithm [Ho 1995] from the package scikit-learn for Python in version 0.23.1. These methods are also compatible with the classifier chain implementation provided by the library sk-multilearn in version 0.2.0.

The hyper parameter optimisation was performed by grid search on suitable search sets. For SVM the parameters X and Y were taken into account, for neural networks, different layer topologies were considered. The parameter choice delivering the best prediction results are X and Y for SVM and a topology consisting of $[n,100,30]$ layers for neural networks, where $n$ represents the number of words in the dictionary used to transform the data.

## 4.2 Classification algorithms for multi-label problems

The prediction of allergens is a multi-label problem because multiple allergens can be assigned to a single product. There are several approaches to solving such problems, two of which are being investigated - binary relevance and classifier chains.

### 4.2.1 Binary relevance

Binary relevance trains an ensemble of binary (yes or no) classifiers, one for each allergen [Burkhardt 2015, Zhang 2018]. Each of these classifiers predicts whether the corresponding allergen is contained in the product or not. The combination of all predictions is then interpreted as a multi-label prediction. This procedure is relatively easy to implement and can be parallelized, but does not take into account possible correlations between the allergens.

### 4.2.2 Classifier chains

If several binary classifiers $C_0, C_1, \ldots, C_n$ are combined in such a way that the classifier $C_i$ uses all predictions of the classifiers $C_j$, with $j < 1$, a classification chain is obtained [Dembczyński 2012, Tenenboim-Chekina 2013, da Silva 2014, Liu 2018]. In this way, the interdependencies of allergens can be taken into account. As many classificators are trained as there are different allergens. In comparison to binary relevance, however, these are sequentially dependent on each other. Direct parallelisation is therefore not possible. The order in which the classifiers are called has an influence on the result. There are strategies how to choose this order as favourable as possible [da Silva 2014]. Fig. 4 shows the principle of a classification chain for the case of three different classifiers.

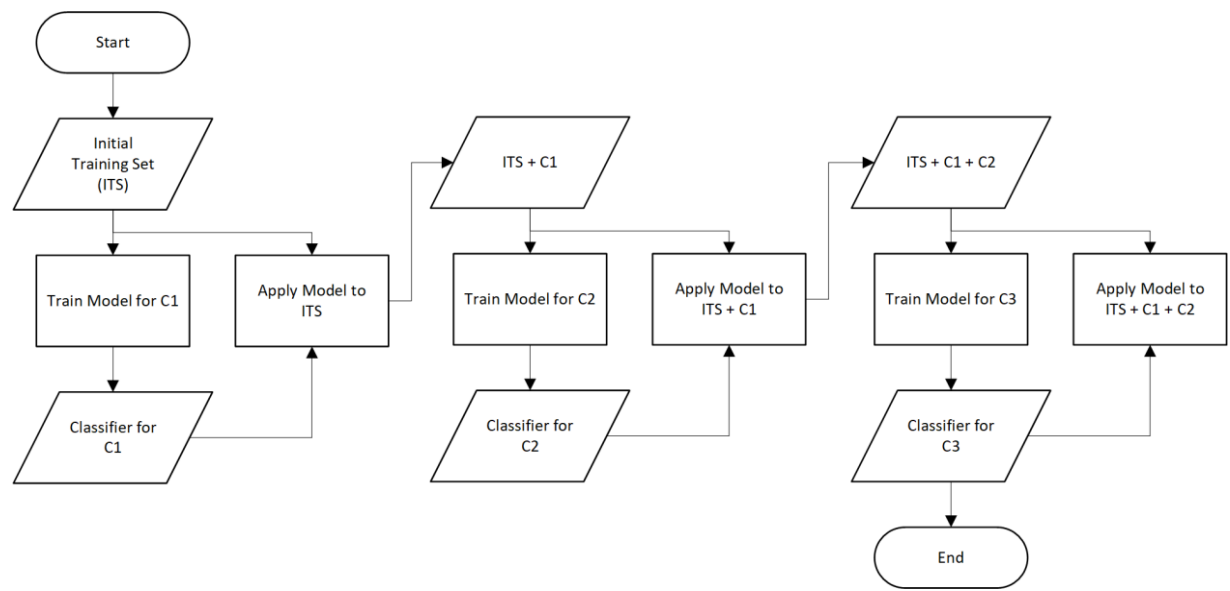

*Fig 4: Schematic depiction of a classifier chain*

### 4.2.3 Metrics

#### 4.2.3.1 Standard metrics

For evaluation purposes we use the standard metrics precision (Pr), recall (Re) and F$_1$-score (F$_1$). Let $Y_x$ be the set of all correct labels and $P_x$ be the set of all predicted labels of a data set $x$. Then $F_x^P$ denotes the set of all false positive labels and $F_x^N$ the set of all false negative labels, whereas $T_x^P$ and $T_x^N$ stand for the sets of all true positive respectively true negative labels.

$$Pr(P_x) = \frac{|T_x^P|}{|T_x^P \cup F_x^P|}, \quad Re(P_x) = \frac{|T_x^P|}{|T_x^P \cup F_x^N|}, \quad F_1(P_x) = \frac{Pr(P_x) \cdot Re(P_x)}{Pr(P_x) + Re(P_x)}$$

#### 4.2.3.2 Alpha evaluation

The alpha evaluation [Boutell 2004] is a metric that allows to weight false positive and false negative predictions differently. The alpha evaluation of a prediction $P_x$ is defined as follows:

$$alpha(P_x) = \left(1 - \frac{\beta |F_x^N| + \gamma |F_x^P|}{|Y_x \cup P_x|}\right)^\alpha, \quad \text{with } \alpha \geq 0 \text{ and } 0 \leq \beta, \gamma \leq 1$$

The so-called forgiveness rate $\alpha$ determines how lenient the metric generally reacts to errors. Small values of α are more aggressive and more forgiving of errors than large values which punish errors more severely. For the borderline case $\alpha = \infty$ the metric becomes 1 only if all predictions are correct. In the opposite case $\alpha = 0$ the metric always takes the value 1 unless all predictions are wrong. For allergen prediction we choose a rather high value of 7.0. The parameter $\beta$ weights the false negative labels, while the parameter $\gamma$ weights the false positive labels. A suitable parameter selection therefore allows us to focus on one of these two wrongly predicted label classes. In the present case of allergen prediction it is less problematic to obtain false positive predictions than false negative ones. If an allergen is assigned to a product that does not actually contain it, this does not represent a health risk for the consumer. However, if a contained allergen is not predicted, the consumption of the product can lead to life-threatening reactions in people susceptible to this allergen. Therefore, in order to assess the prediction quality of the classification algorithms considered, emphasis is placed on the false negative predictions and the parameters are chosen as follows:

$$\alpha = 7.0, \beta = 0.33, \gamma = 1.0$$

#### 4.2.3.3 Multi-label characteristics

The degree of "multi-label-ness" of a data set can be expressed by the two metrics label cardinality ($L_c$) and label density ($L_D$). If $Y_i$ are the labels (specific allergens) of the i$^{th}$ entry (product), then the label cardinality is defined as the average number of labels (specific allergens) per entry (product):

$$L_c = \frac{1}{N} \sum_{i=1}^{N} |Y_i|$$

The label density is defined as the number of labels (specific allergens) per entry (product) divided by the total number of all labels (specific allergens), averaged over all entries (products):

$$L_D = \frac{1}{N} \sum_{i=1}^{N} \frac{|Y_i|}{|L|}, \text{with } L = \bigcup_{i=1}^{N} Y_i$$

For the present data set these metrics have the values $L_c = 5.048$ and $L_D = 0.194$. According to [Silva 2013], label cardinality and density influence the training of classifiers.

# 5 Nutrient consistency verification

We propose a rule-based verification of the FIC-compliant consistency of the nutrient declaration displayed on a product. To this end, the FIC requirements are converted into formula-based rules and applied to each product in the test data set. This allows the definition of defect classes which provide information about the FIC conformity of the products.

## 5.1 Energy value

Three basic verification checks regarding the conformity of the energy value declarations of a product can be directly deduced from the FIC regulation.

### 5.1.1 Total energy

A basic check can be carried out by adding the quantities $m_x$ of a nutrient $x$ multiplied by the corresponding energy values $EV_x$ given in table X:

$$kJ = m_{CH} \cdot EV_{CH} + m_{Prot} \cdot EV_{Prot} + m_{Fat} \cdot EV_{Fat} + m_{Alc} \cdot EV_{Alc} + m_{Fib} \cdot EV_{Fib}$$

The corresponding value in kcal is obtained by multiplying the energy value in kJ by the conversion factor

$$k_{cal}^J = \frac{1}{4,1868} = 0,239$$

In practice, the conversion from kJ to kcal cannot be calculated by multiplication by $k_{cal}^J$, since different conversion factors apply to the individual nutritional values and the total gross energy value is the sum of these (see Article 31: "(1) The energy value shall be calculated using the conversion factors listed in Annex XIV."). The implicit conversion factors calculated in table 1 vary between 4.0 and 4.33 - therefore the conversion factor for the total energy value can never be obtained.

### 5.1.2 Maximum energy value

The total energy value must not be higher than 900 kcal / 100g or 3805 kJ / 100g

### 5.1.3 Energy declaration units of measurement

The energy values for kcal and kJ must be maintained together and cannot stand alone.

## 5.2 Nutrient quantities

### 5.2.1 Single nutrients

The sum of all individual quantities $m_x$ of nutrients $x$ must not exceed 100g (except for ME ml):

$$m_{CH} + m_{Pro} + m_{Fat} + m_{Alc} + m_{Fib} + m_{Sal} \leq 100$$

### 5.2.2 Sub-component quantities

The subcomponent amounts of *un-/saturated fatty acids* for *fat* and *sugar* for *carbohydrates* must not be greater than the basic amounts for *fat* and *sugar*.

$$m_{uFA} + m_{sFA} \leq m_{Fat}$$
$$m_{Sug} \leq m_{CH}$$

## 5.3 Implemented verification checks

From the rules described in sections 5.1-5.3, the error classes described in Table 3 are derived. For each of them, a rule-based test was implemented.

| Error ID | Cause |
|---|---|
| MV_KJ | Declaration of energy [kJ] missing |
| MV_KC | Declaration of energy [kcal] missing |
| CE_EN | Conversion factor [kJ] to [kcal] outside of tolerance bounds |
| SE_EN | Sum of energy values differs too much from total energy value |
| VE_FA | Contains more fatty acids [g] than fat [g] |
| VE_SU | Contains more sugar [g] than carbohydrates [g] |
| VE_IN | Contains more than 100g per 100g of a nutrient |

*Table 5: FIC relevant data errors with assigned error ID*

### 5.3.1 Results using company data

Based on the FIC specifications, the checks implemented in section 5.3 regarding FIC-compliant nutrient information were applied to the company data set presented in section 3.2. The results are shown in Table 6.

| Error ID | MV_KJ | MV_KC | CE_EN | SE_EN | VE_FA | VE_SU | VE_IN |
|---|---|---|---|---|---|---|---|
| #Products | 21949 | 3964 | 1767 | 37521 | 396 | 357 | 12 |

*Table 6: Result of nutrient verification*

As can be clearly seen, in almost half of all cases the information on the energy value in kJ is missing. The corresponding value in kcal, however, is maintained for over 90% of the products. According to FIC, however, the energy value must be indicated for both units. The conversion factor from kJ to kcal is also often outside a tolerance range of 4.1 and 4.3. The exact physical conversion factor of 4.1868 is practically not maintained for any product. This is because different conversion factors are implicitly used for the individual nutrients in FIC, which vary between 4.0 and 4.33. This leads to the paradox that the conversion factor for total energy value, which is actually prescribed, cannot be achieved.

Figures 5 and 6 show the pair-wise error occurrence in mutual dependence in the form of heat maps.

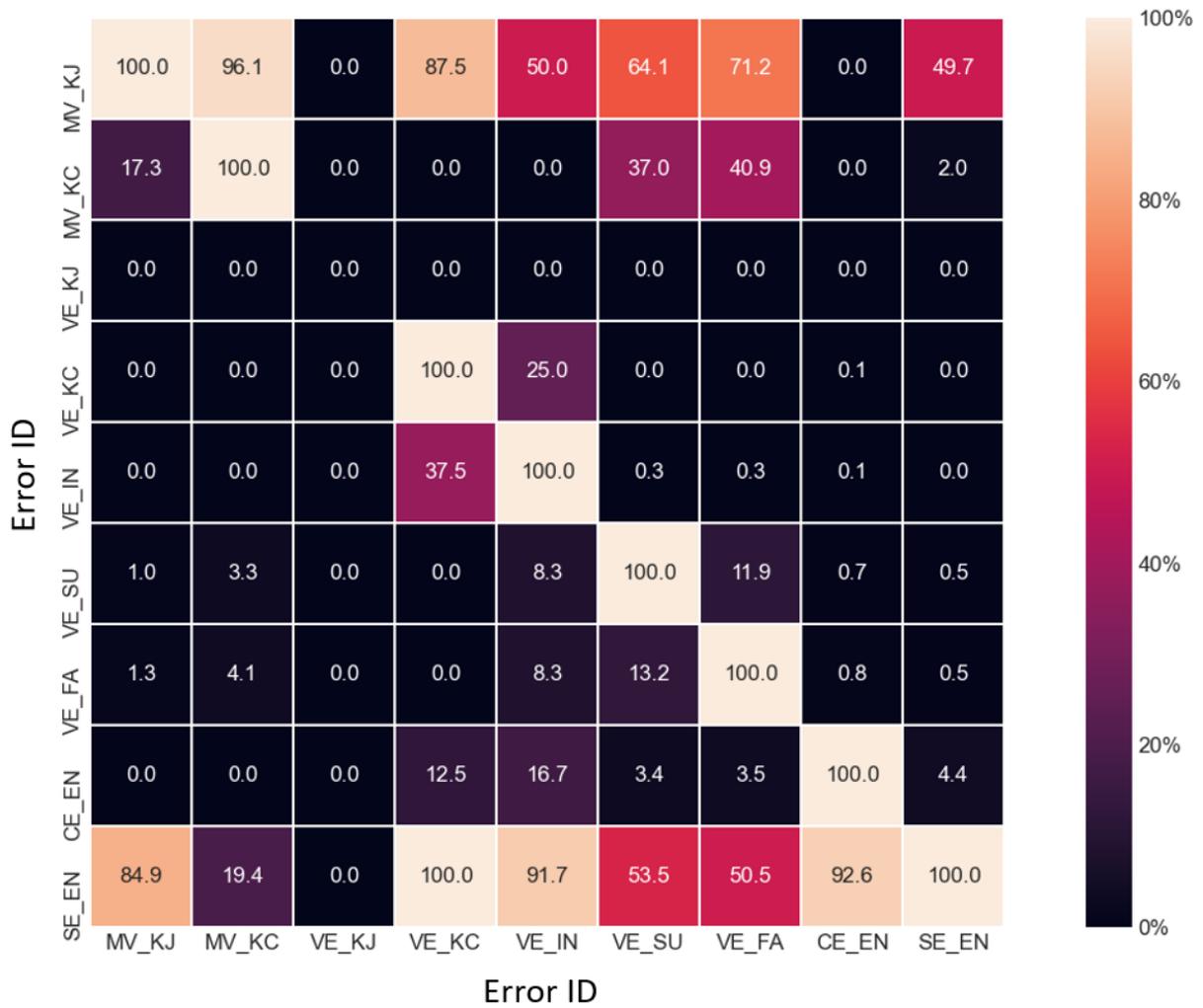

*Fig. 5: Relative frequencies of pairwise error ID occurrence [%]*

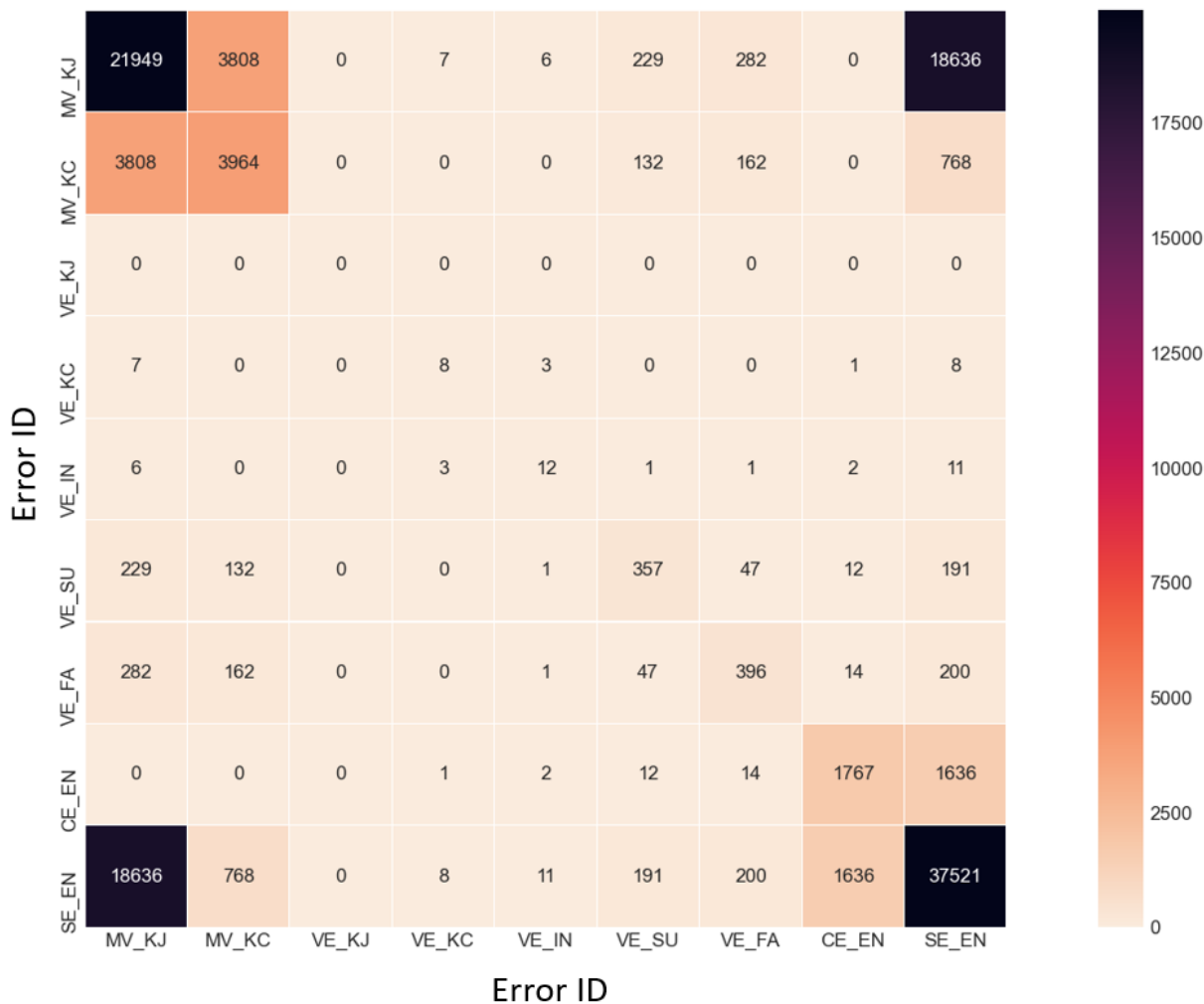

*Fig. 6: Absolute frequencies of pairwise error ID occurrence*

# 6 Allergen prediction using machine learning

## 6.1 Allergen warning verification

All FIC relevant allergens are considered (see 2.2 (Annex II of the Food Information Regulation (FIR) Regulation (EU) No 1169/2011)). In the present product data allergens according to the providing company's specification are also listed. These subdivide the FIC relevant allergens into finer categories. For example, instead of the general allergen fish, a division into the individual fish species is used. There are therefore specific allergens for salmon, trout and other, partly exotic, fish. In the context of the question of FIC consistency of product data, these additional or finer subdivided allergens are not taken into account.

In principle, there are two questions regarding allergens: are there any allergens in a product and if yes, which specific allergens are contained in it? In both cases, the list of ingredients is used for prediction using machine learning methods.

A hybrid approach with a rule-based part is also conceivable. For this purpose, the list of ingredients is checked for signal words that are assigned to corresponding allergens using a predefined dictionary.

Especially for allergens that are only very rarely found in the product data, this hybrid approach can mean a quality gain.

The list of ingredients was first converted to lower case letters for further use, then special characters and numbers were removed. In a separate step, all allergenic ingredients listed in capital letters were extracted and used as an additional optional feature.

## 6.2 Binary Relevance

### 6.2.1 Predicting general allergen incidence

First of all, the classification into products that may or may not contain allergens based on their ingredient list is considered.

The complete list of ingredients is used as a basis, which contains about 40,000 individual ingredients (words). From this, the most frequent words are determined with the help of various parameters for the lower frequency of word occurrence (min_df = 0.01, 0.001, 0.0001, 0.00003). In this way, 244, 1226, 4634 and 9456 ingredients with the highest word occurrence have been taken into account as parameter Voc(abulary). The use of all ingredients increases the calculation time enormously and does not give better results, as tests have shown. Table 8. indicates that a neural network (NN) with 9456 words in bag-of-words (BOW) format gives the best results regarding all considered metrics. Of 2566 allergen-containing products, 2366 are successfully detected, while 5405 allergen-free products are correctly predicted (see section 10.1.1 in the appendix). 42 allergen-free products are incorrectly classified as containing allergens and 200 allergenic products misclassified as free of allergens. The method therefore provides very good results in predicting and elucidating the general allergen incidence of products.

| Algo | Voc | TT | $Pr_{macro}$ | $Re_{macro}$ | $F1_{macro}$ | $Pr_{micro}$ | $Re_{micro}$ | $F1_{micro}$ | Alpha |
|---|---|---|---|---|---|---|---|---|---|
| NN | 244 | BOW | 0.925 | 0.947 | 0.934 | 0.942 | 0.942 | 0.942 | 0.687 |
| NN | 244 | TF-IDF | 0.928 | 0.942 | 0.935 | 0.942 | 0.942 | 0.942 | 0.705 |
| NN | 1226 | BOW | 0.959 | 0.972 | 0.965 | 0.969 | 0.969 | 0.969 | 0.823 |
| NN | 1226 | TF-IDF | 0.954 | 0.971 | 0.962 | 0.967 | 0.967 | 0.967 | 0.804 |
| NN | 4634 | BOW | 0.97 | 0.980 | 0.975 | 0.978 | 0.978 | 0.978 | 0.875 |
| NN | 4634 | TF-IDF | 0.968 | 0.977 | 0.972 | 0.976 | 0.976 | 0.976 | 0.866 |
| NN | 9456 | BOW | **0.977** | **0.983** | **0.980** | **0.982** | **0.982** | **0.982** | **0.903** |
| NN | 9456 | TF-IDF | 0.975 | 0.982 | 0.979 | 0.982 | 0.982 | 0.982 | 0.898 |
| SVM | 244 | BOW | 0.886 | 0.921 | 0.899 | 0.908 | 0.908 | 0.908 | 0.508 |
| SVM | 244 | TF-IDF | 0.89 | 0.922 | 0.902 | 0.911 | 0.911 | 0.911 | 0.530 |
| SVM | 1226 | BOW | 0.941 | 0.960 | 0.949 | 0.956 | 0.956 | 0.956 | 0.755 |
| SVM | 1226 | TF-IDF | 0.945 | 0.961 | 0.952 | 0.957 | 0.957 | 0.957 | 0.761 |
| SVM | 4634 | BOW | 0.962 | 0.970 | 0.966 | 0.970 | 0.970 | 0.970 | 0.842 |
| SVM | 4634 | TF-IDF | 0.959 | 0.968 | 0.963 | 0.968 | 0.968 | 0.968 | 0.834 |
| SVM | 9456 | BOW | 0.957 | 0.964 | 0.961 | 0.966 | 0.966 | 0.966 | 0.826 |
| SVM | 9456 | TF-IDF | 0.961 | 0.966 | 0.964 | 0.968 | 0.968 | 0.968 | 0.842 |
| RF | 244 | BOW | 0.931 | 0.949 | 0.939 | 0.947 | 0.947 | 0.947 | 0.716 |

| RF | 244 | TF-IDF | 0.920 | 0.935 | 0.927 | 0.935 | 0.935 | 0.935 | 0.680 |
| --- | --- | --- | --- | --- | --- | --- | --- | --- | --- |
| RF | 1226 | BOW | 0.957 | 0.967 | 0.962 | 0.966 | 0.966 | 0.966 | 0.819 |
| RF | 1226 | TF-IDF | 0.950 | 0.957 | 0.954 | 0.96 | 0.960 | 0.960 | 0.800 |
| RF | 4634 | BOW | 0.967 | 0.974 | 0.970 | 0.974 | 0.974 | 0.974 | 0.867 |
| RF | 4634 | TF-IDF | 0.963 | 0.963 | 0.963 | 0.968 | 0.968 | 0.968 | 0.855 |
| RF | 9456 | BOW | 0.965 | 0.974 | 0.970 | 0.974 | 0.974 | 0.974 | 0.860 |
| RF | 9456 | TF-IDF | 0.965 | 0.964 | 0.965 | 0.970 | 0.970 | 0.970 | 0.866 |

*Table 8: Prediction results of the presence of allergens using the list of ingredients*

### 6.2.2 Predicting specific allergen incidence

In contrast to 6.2.1, each individual allergen is now to be predicted separately. This is done both by using the entire list of ingredients (6.2.2.1) and by exclusively considering ingredients containing allergens (6.2.2.2).

#### *6.2.2.1 Using all ingredients*

| Allergen | Algo | Voc | TT | $Pr_{macro}$ | $Re_{macro}$ | $F1_{macro}$ | $Pr_{micro}$ | $Re_{micro}$ | $F1_{micro}$ | Alpha |
| --- | --- | --- | --- | --- | --- | --- | --- | --- | --- | --- |
| Milk | NN | 9456 | BOW | 0.968 | 0.970 | 0.969 | 0.969 | 0.969 | 0.969 | 0.880 |
| Soybeans | NN | 9456 | TF-IDF | 0.964 | 0.958 | 0.961 | 0.971 | 0.971 | 0.971 | 0.859 |
| Gluten | NN | 4634 | BOW | 0.942 | 0.949 | 0.946 | 0.953 | 0.953 | 0.953 | 0.810 |
| Sulphur | NN | 4634 | TF-IDF | 0.963 | 0.932 | 0.947 | 0.988 | 0.988 | 0.988 | 0.934 |
| Eggs | NN | 9456 | BOW | 0.967 | 0.963 | 0.965 | 0.974 | 0.974 | 0.974 | 0.874 |
| Fish | NN | 9456 | TF-IDF | 0.969 | 0.933 | 0.950 | 0.987 | 0.987 | 0.987 | 0.929 |
| Nuts | NN | 4634 | TF-IDF | 0.950 | 0.942 | 0.946 | 0.967 | 0.967 | 0.967 | 0.841 |
| Celery | NN | 4634 | BOW | 0.965 | 0.947 | 0.956 | 0.982 | 0.982 | 0.982 | 0.907 |
| Mustard | NN | 1226 | BOW | 0.960 | 0.953 | 0.956 | 0.981 | 0.981 | 0.981 | 0.908 |
| Peanuts | NN | 9456 | BOW | 0.948 | 0.918 | 0.932 | 0.984 | 0.984 | 0.984 | 0.914 |
| Sesame | NN | 1226 | BOW | 0.946 | 0.905 | 0.924 | 0.982 | 0.982 | 0.982 | 0.902 |
| Lupine | NN | 9456 | BOW | 0.903 | 0.859 | 0.880 | 0.991 | 0.991 | 0.991 | 0.953 |
| Molluscs | NN | 9456 | TF-IDF | 0.948 | 0.885 | 0.914 | 0.995 | 0.995 | 0.995 | 0.969 |
| Crustaceans | NN | 9456 | BOW | 0.947 | 0.868 | 0.903 | 0.993 | 0.993 | 0.993 | 0.961 |

*Table 9: Results of the best methods for each allergen using the list of ingredients*

For the selected parameterisation of the alpha metric, neural networks (NN) provide the best results. The Random Forest (RF) algorithm achieves slightly lower metrics while Support Vector Machines (SVM) are inferior to both for the most part (full results can be found in section 10.1.1 in the appendix). The tendency that more extensive dictionaries achieve better results is clearly visible. BOW is more often a better choice of text transformation than TF-IDF. In general, it should be noted that the alpha metric is considered decisive in this evaluation. If one considers Precision, Recall and $F_1$-score, NN and RF are equally good. As the allergens predicted as false negative (FN) are much more critical for the allergen prediction than the false positive (FP) ones due to the potential danger for the consumer, the alpha metric was parameterised accordingly. False negative predictions are weighted three times more than false positive predictions.

### 6.2.2.2 Using allergenic ingredients only

In the present data set, the allergenic ingredients are indicated in capital letters within the list of ingredients. According to FIC, these ingredients must either be printed in bold or appear in capital letters on the product. However, typical ERP systems and databases do not support bold letters. Therefore, it cannot be assumed that the format of the allergenic ingredients can be distinguished from the other ingredients. To compare the prediction based on the whole ingredient list, regular expressions were used to extract the capitalised ingredients and used as input for the models employed in 6.1. Due to the univocal results of 6.2.2.1 only neural networks are considered from now on. Note that the number of words used (Voc) for classifier training is smaller as in 6.2.1, although the same parameter choices for min_df were made. This is due to the fact that the allergenic ingredients constitute only a small subset of the entire set of ingredients and thus fewer words satisfy the required lower frequency of word occurrence given by the parameter min_df.

| Allergen | Algo | Voc | TT | $Pr_{macro}$ | $Re_{macro}$ | $F1_{macro}$ | $Pr_{micro}$ | $Re_{micro}$ | $F1_{micro}$ | Alpha |
|---|---|---|---|---|---|---|---|---|---|---|
| Milk | NN | 1225 | BOW | 0,941 | 0,952 | 0,945 | 0,948 | 0,948 | 0,948 | 0,723 |
| Soybeans | NN | 1225 | BOW | 0,926 | 0,916 | 0,920 | 0,927 | 0,927 | 0,927 | 0,651 |
| Gluten | NN | 3432 | BOW | 0,926 | 0,926 | 0,926 | 0,927 | 0,927 | 0,927 | 0,693 |
| Sulphur | NN | 1225 | TFIDF | 0,971 | 0,934 | 0,952 | 0,985 | 0,985 | 0,985 | 0,916 |
| Eggs | NN | 1225 | BOW | 0,946 | 0,932 | 0,938 | 0,944 | 0,944 | 0,944 | 0,707 |
| Fish | NN | 3432 | TFIDF | 0,953 | 0,920 | 0,936 | 0,978 | 0,978 | 0,978 | 0,880 |
| Nuts | NN | 1225 | BOW | 0,921 | 0,911 | 0,916 | 0,934 | 0,934 | 0,934 | 0,694 |
| Celery | NN | 1225 | BOW | 0,953 | 0,903 | 0,926 | 0,961 | 0,961 | 0,961 | 0,780 |
| Mustard | NN | 1225 | TFIDF | 0,963 | 0,924 | 0,942 | 0,966 | 0,966 | 0,966 | 0,804 |
| Peanuts | NN | 1225 | TFIDF | 0,911 | 0,858 | 0,882 | 0,961 | 0,961 | 0,961 | 0,799 |
| Sesame | NN | 1225 | BOW | 0,937 | 0,857 | 0,892 | 0,963 | 0,963 | 0,963 | 0,792 |
| Lupine | NN | 1225 | TFIDF | 0,909 | 0,792 | 0,84 | 0,986 | 0,986 | 0,986 | 0,917 |
| Molluscs | NN | 3432 | TFIDF | 0,842 | 0,801 | 0,82 | 0,987 | 0,987 | 0,987 | 0,935 |
| Crustaceans | NN | 3432 | BOW | 0,906 | 0,824 | 0,86 | 0,984 | 0,984 | 0,984 | 0,913 |

*Table 10: Results of the best methods for each allergen using allergenic ingredients only*

Comparing the results shown in table 10 with those using the complete list of ingredients (table 9), it is clear that using the complete list for each specific allergen provides better predictions than restricting the list of ingredients to those containing only allergens. Although the underlying dictionaries are more comprehensive than in the purely allergenic case, even for sesame, for which there is almost the same amount of words in both models 1225 and 1226 respectively, the complete ingredient list provides a significantly higher alpha metric of 0.902 than the purely allergenic ingredient list with 0.792. This could be due to combinations of non-allergenic ingredients hinting more strongly at allergens, than using only the allergenic ingredients themselves - which are usually not available in the data sets anyway.

## 6.3 Classifier Chains

Prediction with chained classifiers depends strongly on the order in which the individual allergen-specific classifiers are called. Therefore, for each combination of text transformation and vocabulary size, ten or twenty runs with randomly permuted order were performed and the metrics averaged. For comparison purposes, an optimised sequence was determined on the basis of the results from Table 9 with regard to the mutual dependence of the allergens (see Fig. 3) and the corresponding classifier chain was also calculated for this sequence.

### 6.3.1 Predicting specific allergen incidence (averaged randomized order)

Text transformation BOW and 4634 words, averaged over 10 runs with randomized order.

| Allergen | Voc | $Pr_{macro}$ | $Re_{macro}$ | $F1_{macro}$ | $Pr_{micro}$ | $Re_{micro}$ | $F1_{micro}$ | Alpha |
|---|---|---|---|---|---|---|---|---|
| Milk | 4634 | 0,525 | 0,333 | 0,244 | 0,576 | 0,576 | 0,576 | 0,000 |
| Soybeans | 4634 | 0,422 | 0,345 | 0,315 | 0,749 | 0,749 | 0,749 | 0,065 |
| Gluten | 4634 | 0,317 | 0,256 | 0,220 | 0,682 | 0,682 | 0,682 | 0,014 |
| Sulphur | 4634 | 0,313 | 0,332 | 0,322 | 0,936 | 0,936 | 0,936 | 0,620 |
| Eggs | 4634 | 0,585 | 0,334 | 0,287 | 0,755 | 0,755 | 0,755 | 0,065 |
| Fish | 4634 | 0,324 | 0,333 | 0,321 | 0,925 | 0,925 | 0,925 | 0,563 |
| Nuts | 4634 | 0,288 | 0,333 | 0,309 | 0,865 | 0,865 | 0,865 | 0,305 |
| Celery | 4634 | 0,233 | 0,250 | 0,241 | 0,932 | 0,932 | 0,932 | 0,591 |
| Mustard | 4634 | 0,309 | 0,333 | 0,321 | 0,928 | 0,928 | 0,928 | 0,571 |
| Peanuts | 4634 | 0,327 | 0,333 | 0,330 | 0,981 | 0,981 | 0,981 | 0,874 |
| Sesame | 4634 | 0,404 | 0,493 | 0,381 | 0,794 | 0,794 | 0,794 | 0,529 |
| Lupine | 4634 | 0,387 | 0,443 | 0,380 | 0,855 | 0,855 | 0,855 | 0,623 |
| Molluscs | 4634 | 0,525 | 0,333 | 0,244 | 0,576 | 0,576 | 0,576 | 0,000 |
| Crustaceans | 4634 | 0,422 | 0,345 | 0,315 | 0,749 | 0,749 | 0,749 | 0,065 |

*Table 11: Results using classifier chains and ingredient list with randomized order and 10x averaging*

Text transformation BOW and 4634 words, averaged over 10 runs with randomized order.

| Allergen | Voc | $Pr_{macro}$ | $Re_{macro}$ | $F1_{macro}$ | $Pr_{micro}$ | $Re_{micro}$ | $F1_{micro}$ | Alpha |
|---|---|---|---|---|---|---|---|---|
| Milk | 1226 | 0,374 | 0,334 | 0,245 | 0,576 | 0,576 | 0,576 | 0,000 |
| Soybeans | 1226 | 0,226 | 0,248 | 0,217 | 0,737 | 0,737 | 0,737 | 0,053 |
| Gluten | 1226 | 0,298 | 0,332 | 0,272 | 0,674 | 0,674 | 0,674 | 0,011 |
| Sulphur | 1226 | 0,240 | 0,246 | 0,242 | 0,918 | 0,918 | 0,918 | 0,587 |
| Eggs | 1226 | 0,436 | 0,347 | 0,319 | 0,757 | 0,757 | 0,757 | 0,074 |
| Fish | 1226 | 0,309 | 0,332 | 0,320 | 0,925 | 0,925 | 0,925 | 0,562 |
| Nuts | 1226 | 0,453 | 0,254 | 0,231 | 0,811 | 0,811 | 0,811 | 0,165 |
| Celery | 1226 | 0,354 | 0,334 | 0,313 | 0,880 | 0,880 | 0,880 | 0,363 |
| Mustard | 1226 | 0,420 | 0,414 | 0,409 | 0,804 | 0,804 | 0,804 | 0,361 |
| Peanuts | 1226 | 0,374 | 0,406 | 0,373 | 0,845 | 0,845 | 0,845 | 0,497 |
| Sesame | 1226 | 0,384 | 0,434 | 0,381 | 0,810 | 0,810 | 0,810 | 0,449 |

| Lupine | 1226 | 0,348 | 0,438 | 0,329 | 0,814 | 0,814 | 0,814 | 0,558 |
|---|---|---|---|---|---|---|---|---|
| Molluscs | 1226 | 0,413 | 0,511 | 0,391 | 0,809 | 0,809 | 0,809 | 0,552 |
| Crustaceans | 1226 | 0,404 | 0,473 | 0,397 | 0,851 | 0,851 | 0,851 | 0,610 |

*Table 12: Results using classifier chains and ingredient list with randomized order and 20x averaging*

### 6.3.2 Predicting specific allergen incidence (optimized order)

The advantage of classifier chains lies their capability to consider dependencies between the individual labels. Based on Fig. 2, an optimised sequence was therefore determined, attempting to model the individual allergens in their presumed dependencies on each other. Starting with the allergen gluten, which is the least dependent on other allergens, this optimised sequence is determined in the order of the highest dependency:

Gluten → Molluscs → Crustaceans → Lupine → Sesame → Peanuts → Sulphur → Celery → Eggs → Nuts → Milk → Fish → Mustard → Soybeans

| Allergen | $Pr_{macro}$ | $Re_{macro}$ | $F1_{macro}$ | $Pr_{micro}$ | $Re_{micro}$ | $F1_{micro}$ | Alpha |
|---|---|---|---|---|---|---|---|
| Milk | 0,359 | 0,333 | 0,244 | 0,576 | 0,576 | 0,576 | 0,000 |
| Soybeans | 0,632 | 0,622 | 0,627 | 0,956 | 0,956 | 0,956 | 0,780 |
| Gluten | 0,226 | 0,333 | 0,269 | 0,678 | 0,678 | 0,678 | 0,011 |
| Sulphur | 0,313 | 0,333 | 0,323 | 0,937 | 0,937 | 0,937 | 0,623 |
| Eggs | 0,363 | 0,333 | 0,287 | 0,754 | 0,754 | 0,754 | 0,065 |
| Fish | 0,309 | 0,333 | 0,321 | 0,927 | 0,927 | 0,927 | 0,567 |
| Nuts | 0,162 | 0,200 | 0,179 | 0,808 | 0,808 | 0,808 | 0,159 |
| Celery | 0,294 | 0,333 | 0,312 | 0,881 | 0,881 | 0,881 | 0,363 |
| Mustard | 0,360 | 0,383 | 0,350 | 0,666 | 0,666 | 0,666 | 0,146 |
| Peanuts | 0,341 | 0,334 | 0,324 | 0,930 | 0,930 | 0,930 | 0,588 |
| Sesame | 0,309 | 0,332 | 0,320 | 0,926 | 0,926 | 0,926 | 0,567 |
| Lupine | 0,327 | 0,333 | 0,330 | 0,981 | 0,981 | 0,981 | 0,874 |
| Molluscs | 0,333 | 0,333 | 0,333 | 0,960 | 0,960 | 0,960 | 0,836 |
| Crustaceans | 0,250 | 0,250 | 0,250 | 0,956 | 0,956 | 0,956 | 0,802 |

*Table 13: Results using classifier chains and ingredient list with optimized order*

The results show that the prediction of allergens using classifier chains gives significantly worse results than using the binary relevance approach. This could be due to the training data, which is extended by one feature with each link in the chain, thus distracting the resulting model too much from the list of ingredients and instead focusing on the already predicted allergens. Implicitly this means that the specific allergens are independent of each other, otherwise the additional information that the classification chain exploits should lead to a better prediction. There is no quality gain between randomly chosen order and optimised order of allergens. This also indicates that the individual allergens are independent of each other.

# 7  Conclusion

The rule-based testing of nutrients for FIC compliance has revealed two distinct problems in the test data. Firstly, the energy value data in kJ and kcal are maintained in clear contradiction to the FIC specifications. Nearly half of the products display only the kcal value and not both kcal and kJ values.

Secondly, consistency of the energy values is not given. The conversion factor from kcal to kJ given in the FIC is rarely used in the energy value specification. This is due to the fact that the conversion factors for each basic component are defined differently in the FIC to avoid floating point numbers. The sum of the individual basic constituents cannot therefore satisfy the actually also prescribed conversion factor of 4.1868.

The checks implemented within the scope of this report can be easily extended to portion and daily dose declarations. The developed data model allows the use of the presented rule and ML-based checks for data from various sources by means of mapping tables. The integration of further checks is made possible by the comprehensive modelling of all FIC-relevant attributes

Allergen prediction by machine learning gives good results with some variation depending on the allergen. This is also due to the fact that allergens are not evenly distributed in the data, but rarer allergens such as lupine or molluscs are only present in a few products, compared to gluten or milk, for example. This reduces the accuracy of the prediction. It has been shown that the binary relevance approach gives better results than classifier chains. This is due to the fact that the occurrence of allergens is largely independent of each other. Some combinations such as gluten and milk are dependent to a certain extent, but there is no pairwise interdependence across the whole allergen spectrum. This is why the assumption of dependence distorts the prediction and leads to poor results. Classifier chains, which require more computing time, can therefore be excluded in favour of the more parallelizable binary relevance algorithms.

To further increase the prediction quality, ensembles of several different models can be examined or the topology of the neural network used can be further optimised. Moreover, additional features such as product name or brand could further improve the results.

# 8  Acknowledgements

This work is part of the KMU-innovativ project "Intelligent Master Data Quality Assurance Assistant" (IMQAA) funded by the German Ministry of Education and Research under grant number 01|S18018.

# 9   References

[Boutell 2004] Boutell, Matthew R.; Luo, Jiebo; Shen, Xipeng; Brown, Christopher M. (2004): Learning multi-label scene classification, *Pattern Recognition*, Vol 37.9, Pages 1757-1771

[Burkhardt 2015] Burkhardt, Sophie; Kramer, Stefan (2015): On the spectrum between binary relevance and classifier chains in multi-label classification. *Proceedings of the 30th Annual ACM Symposium on Applied Computing*, S. 885–892.

[Cortes 1995] Cortes, C. and V. N. Vapnik (1995). Support-vector networks. *Machine Learning*. 20 (3): 273–297

[da Silva 2014] da Silva, Pablo Nascimento; Gonçalves, Eduardo Corrêa; Plastino, Alexandre; Freitas, Alex A. (2014): Distinct Chains for Different Instances: An Effective Strategy for Multi-label Classifier Chains, *Proceedings of the ECML PKDD 2014*, S. 453–468.

[Dembczyński 2012] Dembczyński, Krzysztof; Waegeman, Willem; Hüllermeier, Eyke (2012): An Analysis of Chaining in Multi-Label Classification.

[EU 2011] Regulation (EU) No 1169/2011 of the European Parliament and of the Council of 25 October 2011 on the provision of food information to consumers, amending Regulations (EC) No 1924/2006 and (EC) No 1925/2006 of the European Parliament and of the Council, and repealing Commission Directive 87/250/EEC, Council Directive 90/496/EEC, Commission Directive 1999/10/EC, Directive 2000/13/EC of the European Parliament and of the Council, Commission Directives 2002/67/EC and 2008/5/EC and Commission Regulation (EC) No 608/2004 (Text with EEA relevance), *OJ L 304,* 18–63

[Ho 1995] Ho, T.K. (1995). Random Decision Forests. *Proceedings of the 3rd ICDAR*, 278–282

[Jones 1972] Jones, K. S. (1972). A statistical interpretation of term specificity and its application in retrieval. *Journal of documentation*, 28(1), 11–21

[Liu 2019] Liu, Bin; Tsoumakas, Grigorios (2018): Making Classifier Chains Resilient to Class Imbalance. Online https://www.cs.waikato.ac.nz/~eibe/pubs/ccformlc.pdf.

[Silva 2013] Silva, R., Bernardini, F. (2013). Analyzing the Influence of Cardinality and Density Characteristics on Multi-Label Learning, *Proceedings of BRACIS 2013*

[Schmidthuber 2015] Schmidthuber, J. (2015): Deep learning in neural networks: An overview, *Neural Networks*, Vol 61, 85-117

[Szymański 2017] Szymański, Piotr; Kajdanowicz, Tomasz (2017): A scikit-based Python environment for performing multi-label classification. Online verfügbar unter http://citeseerx.ist.psu.edu/viewdoc/download?doi=10.1.1.407.9705&rep=rep1&type=pdf.

[Tenenboim-Chekina 2013] Tenenboim-Chekina, Lena; Rokach, Lior; Shapira, Bracha (2013): Ensemble of Feature Chains for Anomaly Detection*, Proceedings of the 11th International Workshop, MCS*, S. 295–306.

[Tsoumakas 2007] Tsoumakas, Grigorios; Katakis, Ioannis (2007): Multi-Label Classification - An Overview. In: *International Journal of Data Warehousing and Mining* 3 (3), S. 1–13. DOI: 10.4018/jdwm.2007070101.

[Zhang 2018] Zhang, Min-Ling; Li, Yu-Kun; Liu, Xu-Ying; Geng, Xin (2018): Binary relevance for multi-label learning: an overview. In: *Front. Comput. Sci.* 12 (2), S. 191–202. DOI: 10.1007/s11704-017-7031-7.

[Zhang 2010] Zhang, Min-Ling; Zhang, Kun (2010): Multi-label learning by exploiting label dependency, *Proceedings of the 16th ACM SIGKDD*, S. 999.

[Zhang 2014] Zhang, Min-Ling; Zhou, Zhi-Hua (2014): A Review on Multi-Label Learning Algorithms. In: *IEEE Trans. Knowl. Data Eng.* 26 (8), S. 1819–1837. DOI: 10.1109/TKDE.2013.39

# 10 Appendix

The appendix contains detailed tables of the results of the allergen classification experiments. The best metric values achieved are displayed in bold type.

## 10.1 Results Binary Relevance

### 10.1.1 General allergen occurrence

| Algo | Vocab | TextT | TP | TN | FP | FN | Pr | Re | F1 | Alpha |
|---|---|---|---|---|---|---|---|---|---|---|
| NN | 244 | BOW | 5166 | 2385 | 102 | 360 | 0.946 | 0.942 | 0.943 | 0.687 |
| NN | 244 | TF-IDF | 5093 | 2454 | 146 | 320 | 0.944 | 0.942 | 0.942 | 0.705 |
| NN | 1226 | BOW | 5275 | 2490 | 52 | 196 | 0.97 | 0.969 | 0.969 | 0.823 |
| NN | 1226 | TF-IDF | 5306 | 2439 | 45 | 223 | 0.968 | 0.967 | 0.967 | 0.804 |
| NN | 4634 | BOW | 5380 | 2458 | 41 | 134 | 0.979 | 0.978 | 0.978 | 0.875 |
| NN | 4634 | TF-IDF | 5362 | 2456 | 53 | 142 | 0.976 | 0.976 | 0.976 | 0.866 |
| NN | 9456 | BOW | 5405 | 2466 | 42 | 100 | **0.983** | **0.982** | **0.982** | **0.903** |
| NN | 9456 | TF-IDF | 5405 | 2460 | 41 | 107 | 0.982 | **0.982** | **0.982** | 0.898 |
| SVM | 244 | BOW | 2446 | 4826 | 105 | 636 | 0.920 | 0.908 | 0.910 | 0.508 |
| SVM | 244 | TF-IDF | 2424 | 4875 | 121 | 593 | 0.921 | 0.911 | 0.913 | 0.530 |
| SVM | 1226 | BOW | 2360 | 5302 | 73 | 278 | 0.959 | 0.956 | 0.957 | 0.755 |
| SVM | 1226 | TF-IDF | 2533 | 5135 | 76 | 269 | 0.959 | 0.957 | 0.957 | 0.761 |
| SVM | 4634 | BOW | 2512 | 5262 | 75 | 164 | 0.971 | 0.970 | 0.970 | 0.842 |
| SVM | 4634 | TF-IDF | 2439 | 5319 | 84 | 171 | 0.969 | 0.968 | 0.968 | 0.834 |
| SVM | 9456 | BOW | 2410 | 5329 | 98 | 176 | 0.966 | 0.966 | 0.966 | 0.826 |
| SVM | 9456 | TF-IDF | 2473 | 5284 | 101 | 155 | 0.968 | 0.968 | 0.968 | 0.842 |
| RF | 244 | BOW | 5187 | 2398 | 112 | 316 | 0.949 | 0.947 | 0.947 | 0.716 |
| RF | 244 | TF-IDF | 5123 | 2373 | 172 | 345 | 0.938 | 0.935 | 0.936 | 0.680 |
| RF | 1226 | BOW | 5266 | 2477 | 78 | 192 | 0.967 | 0.966 | 0.966 | 0.819 |
| RF | 1226 | TF-IDF | 5291 | 2398 | 123 | 201 | 0.960 | 0.960 | 0.960 | 0.800 |
| RF | 4634 | BOW | 5370 | 2438 | 70 | 135 | 0.975 | 0.974 | 0.975 | 0.867 |
| RF | 4634 | TF-IDF | 5322 | 2434 | 127 | 130 | 0.968 | 0.968 | 0.968 | 0.855 |
| RF | 9456 | BOW | 5423 | 2382 | 62 | 146 | 0.975 | 0.974 | 0.974 | 0.860 |
| RF | 9456 | TF-IDF | 5381 | 2388 | 128 | 116 | 0.970 | 0.970 | 0.970 | 0.687 |

## 10.1.2 Specific allergen occurrence

## 10.1.3 Milk and lactose

### *10.1.3.1 Using list of ingredients*

| Algo | Vocab | TextT | TP | TN | FP | FN | Pr | Re | F1 | Alpha |
|------|-------|-------|------|------|-----|-----|-------|-------|-------|-------|
| NN | 244 | BOW | 3210 | 4438 | 132 | 233 | 0.955 | 0.954 | 0.954 | 0.773 |
| NN | 244 | TF-IDF | 3276 | 4343 | 165 | 229 | 0.951 | 0.951 | 0.951 | 0.767 |
| NN | 1226 | BOW | 3339 | 4432 | 105 | 137 | 0.970 | 0.970 | 0.970 | 0.855 |
| NN | 1226 | TF-IDF | 3243 | 4540 | 112 | 118 | 0.971 | 0.971 | 0.971 | 0.869 |
| NN | 4634 | BOW | 3327 | 4426 | 112 | 148 | 0.968 | 0.968 | 0.968 | 0.844 |
| NN | 4634 | TF-IDF | 3316 | 4458 | 130 | 109 | 0.970 | 0.970 | 0.970 | 0.871 |
| NN | 9456 | BOW | 3361 | 4405 | 158 | 89 | 0.969 | 0.969 | 0.969 | **0.880** |
| NN | 9456 | TF-IDF | 3396 | 4378 | 103 | 136 | 0.970 | 0.970 | 0.970 | 0.857 |
| SVM | 244 | BOW | 4426 | 3085 | 121 | 381 | 0.939 | 0.937 | 0.937 | 0.668 |
| SVM | 244 | TF-IDF | 4428 | 3087 | 132 | 366 | 0.939 | 0.938 | 0.938 | 0.676 |
| SVM | 1226 | BOW | 4490 | 3217 | 108 | 198 | 0.962 | 0.962 | 0.962 | 0.806 |
| SVM | 1226 | TF-IDF | 4512 | 3185 | 115 | 201 | 0.961 | 0.961 | 0.960 | 0.802 |
| SVM | 4634 | BOW | 4417 | 3203 | 190 | 203 | 0.951 | 0.951 | 0.951 | 0.780 |
| SVM | 4634 | TF-IDF | 4449 | 3202 | 156 | 206 | 0.955 | 0.955 | 0.955 | 0.787 |
| SVM | 9456 | BOW | 4370 | 3165 | 241 | 237 | 0.940 | 0.940 | 0.940 | 0.741 |
| SVM | 9456 | TF-IDF | 4439 | 3212 | 135 | 227 | 0.955 | 0.955 | 0.955 | 0.776 |
| RF | 244 | BOW | 3256 | 4423 | 105 | 229 | 0.959 | 0.958 | 0.958 | 0.783 |
| RF | 244 | TF-IDF | 3233 | 4407 | 148 | 225 | 0.953 | 0.953 | 0.953 | 0.775 |
| RF | 1226 | BOW | 3322 | 4437 | 117 | 137 | 0.968 | 0.968 | 0.968 | 0.852 |
| RF | 1226 | TF-IDF | 3289 | 4429 | 157 | 138 | 0.963 | 0.963 | 0.963 | 0.840 |
| RF | 4634 | BOW | 3363 | 4403 | 110 | 137 | 0.969 | 0.969 | 0.969 | 0.854 |
| RF | 4634 | TF-IDF | 3333 | 4395 | 136 | 149 | 0.964 | 0.964 | 0.964 | 0.837 |
| RF | 9456 | BOW | 3354 | 4432 | 99 | 128 | **0.972** | **0.972** | **0.972** | 0.864 |
| RF | 9456 | TF-IDF | 3346 | 4417 | 135 | 115 | 0.969 | 0.969 | 0.969 | 0.865 |

### 10.1.3.2 Using allergenic ingredients only

| Algo | Vocab | TextT | TP | TN | FP | FN | Pr | Re | F1 | Alpha |
|---|---|---|---|---|---|---|---|---|---|---|
| NN | 75 | BOW | 3054 | 1929 | 94 | 416 | 0.916 | 0.907 | 0.908 | 0.518 |
| NN | 75 | TF-IDF | 3041 | 1982 | 93 | 377 | 0.921 | 0.914 | 0.915 | 0.553 |
| NN | 298 | BOW | 3146 | 1998 | 91 | 258 | 0.939 | 0.936 | 0.937 | 0.668 |
| NN | 298 | TF-IDF | 3178 | 1996 | 92 | 227 | 0.944 | 0.942 | 0.942 | 0.700 |
| NN | 1225 | BOW | 3249 | 1961 | 71 | 212 | **0.950** | **0.948** | **0.949** | **0.723** |
| NN | 1225 | TF-IDF | 3231 | 1942 | 90 | 230 | 0.944 | 0.942 | 0.942 | 0.697 |
| NN | 3432 | BOW | 3251 | 1956 | 65 | 221 | **0.950** | **0.948** | 0.948 | 0.716 |
| NN | 3432 | TF-IDF | 3177 | 1997 | 91 | 228 | 0.944 | 0.942 | 0.942 | 0.699 |

## 10.1.4 Soybeans

### 10.1.4.1 Using list of ingredients

| Algo | Vocab | TextT | TP | TN | FP | FN | Pr | Re | F1 | Alpha |
|---|---|---|---|---|---|---|---|---|---|---|
| NN | 244 | BOW | 1849 | 5851 | 155 | 158 | 0.961 | 0.961 | 0.961 | 0.825 |
| NN | 244 | TF-IDF | 1776 | 5898 | 116 | 223 | 0.957 | 0.958 | 0.957 | 0.785 |
| NN | 1226 | BOW | 1824 | 5922 | 84 | 183 | 0.967 | 0.967 | 0.966 | 0.824 |
| NN | 1226 | TF-IDF | 1830 | 5906 | 106 | 171 | 0.965 | 0.965 | 0.965 | 0.828 |
| NN | 4634 | BOW | 1880 | 5916 | 70 | 147 | **0.973** | **0.973** | 0.973 | 0.857 |
| NN | 4634 | TF-IDF | 1835 | 5873 | 153 | 152 | 0.962 | 0.962 | 0.962 | 0.830 |
| NN | 9456 | BOW | 1891 | 5871 | 127 | 124 | 0.969 | 0.969 | 0.969 | 0.860 |
| NN | 9456 | TF-IDF | 1867 | 5913 | 98 | 135 | 0.971 | 0.971 | **0.971** | **0.859** |
| SVM | 244 | BOW | 5936 | 1602 | 101 | 374 | 0.941 | 0.941 | 0.939 | 0.678 |
| SVM | 244 | TF-IDF | 5887 | 1650 | 101 | 375 | 0.941 | 0.941 | 0.939 | 0.677 |
| SVM | 1226 | BOW | 5895 | 1797 | 109 | 212 | 0.960 | 0.960 | 0.960 | 0.795 |
| SVM | 1226 | TF-IDF | 5833 | 1812 | 150 | 218 | 0.954 | 0.954 | 0.954 | 0.779 |
| SVM | 4634 | BOW | 5788 | 1738 | 233 | 254 | 0.939 | 0.939 | 0.939 | 0.730 |
| SVM | 4634 | TF-IDF | 5856 | 1754 | 176 | 227 | 0.949 | 0.950 | 0.949 | 0.765 |
| SVM | 9456 | BOW | 5754 | 1737 | 246 | 276 | 0.935 | 0.935 | 0.935 | 0.710 |
| SVM | 9456 | TF-IDF | 5820 | 1764 | 175 | 254 | 0.946 | 0.946 | 0.946 | 0.745 |
| RF | 244 | BOW | 1819 | 5902 | 84 | 208 | 0.963 | 0.964 | 0.963 | 0.805 |
| RF | 244 | TF-IDF | 1822 | 5856 | 125 | 210 | 0.958 | 0.958 | 0.958 | 0.792 |

| Algo | Vocab | TextT | TP | TN | FP | FN | Pr | Re | F1 | Alpha |
|---|---|---|---|---|---|---|---|---|---|---|
| RF | 1226 | BOW | 1879 | 5894 | 80 | 160 | 0.970 | 0.970 | 0.970 | 0.844 |
| RF | 1226 | TF-IDF | 1865 | 5860 | 97 | 191 | 0.964 | 0.964 | 0.964 | 0.815 |
| RF | 4634 | BOW | 1821 | 5937 | 86 | 169 | 0.968 | 0.968 | 0.968 | 0.835 |
| RF | 4634 | TF-IDF | 1859 | 5862 | 106 | 186 | 0.963 | 0.964 | 0.963 | 0.816 |
| RF | 9456 | BOW | 1867 | 5863 | 91 | 192 | 0.964 | 0.965 | 0.964 | 0.815 |
| RF | 9456 | TF-IDF | 1763 | 5939 | 107 | 204 | 0.961 | 0.961 | 0.961 | 0.802 |

### *10.1.4.2 Using allergenic ingredients only*

| Algo | Vocab | TextT | TP | TN | FP | FN | Pr | Re | F1 | Alpha |
|---|---|---|---|---|---|---|---|---|---|---|
| NN | 75 | BOW | 1642 | 3389 | 124 | 338 | 0.917 | 0.916 | 0.915 | 0.578 |
| NN | 75 | TF-IDF | 1648 | 3347 | 137 | 361 | 0.910 | 0.909 | 0.908 | 0.552 |
| NN | 298 | BOW | 1742 | 3318 | 157 | 276 | 0.921 | 0.921 | 0.921 | 0.626 |
| NN | 298 | TF-IDF | 1745 | 3333 | 122 | 293 | 0.925 | 0.924 | 0.924 | 0.622 |
| NN | 1225 | BOW | 1766 | 3324 | 149 | 254 | 0.927 | 0.927 | 0.926 | **0.651** |
| NN | 1225 | TF-IDF | 1781 | 3332 | 96 | 284 | **0.932** | **0.931** | **0.930** | 0.640 |
| NN | 3432 | BOW | 1751 | 3352 | 124 | 266 | 0.929 | 0.929 | 0.928 | 0.648 |
| NN | 3432 | TF-IDF | 1726 | 3376 | 121 | 270 | 0.929 | 0.929 | 0.928 | 0.645 |

### 10.1.5 Gluten

### *10.1.5.1 Using list of ingredients*

| Algo | Vocab | TextT | TP | TN | FP | FN | Pr | Re | F1 | Alpha |
|---|---|---|---|---|---|---|---|---|---|---|
| NN | 244 | BOW | 2274 | 5271 | 263 | 205 | 0.942 | 0.942 | 0.942 | 0.759 |
| NN | 244 | TF-IDF | 2390 | 5145 | 248 | 230 | 0.940 | 0.940 | 0.940 | 0.744 |
| NN | 1226 | BOW | 2376 | 5220 | 222 | 195 | 0.948 | 0.948 | 0.948 | 0.777 |
| NN | 1226 | TF-IDF | 2332 | 5264 | 202 | 215 | 0.948 | 0.948 | 0.948 | 0.768 |
| NN | 4634 | BOW | 2382 | 5252 | 227 | 152 | 0.953 | 0.953 | 0.953 | **0.810** |
| NN | 4634 | TF-IDF | 2379 | 5233 | 224 | 177 | 0.950 | 0.950 | 0.950 | 0.791 |
| NN | 9456 | BOW | 2412 | 5230 | 192 | 179 | **0.954** | **0.954** | **0.954** | 0.798 |
| NN | 9456 | TF-IDF | 2366 | 5257 | 188 | 202 | 0.951 | 0.951 | 0.951 | 0.781 |
| SVM | 244 | BOW | 5083 | 2296 | 334 | 300 | 0.921 | 0.921 | 0.921 | 0.670 |
| SVM | 244 | TF-IDF | 5103 | 2314 | 316 | 280 | 0.926 | 0.926 | 0.926 | 0.689 |
| SVM | 1226 | BOW | 5180 | 2337 | 264 | 232 | 0.938 | 0.938 | 0.938 | 0.738 |
| SVM | 1226 | TF-IDF | 5213 | 2315 | 239 | 246 | 0.939 | 0.939 | 0.939 | 0.734 |
| SVM | 4634 | BOW | 5152 | 2266 | 311 | 284 | 0.926 | 0.926 | 0.926 | 0.687 |

| Algo | Vocab | TextT | TP | TN | FP | FN | Pr | Re | F1 | Alpha |
|---|---|---|---|---|---|---|---|---|---|---|
| SVM | 4634 | TF-IDF | 5190 | 2244 | 281 | 298 | 0.928 | 0.928 | 0.928 | 0.685 |
| SVM | 9456 | BOW | 5075 | 2268 | 356 | 314 | 0.917 | 0.916 | 0.917 | 0.654 |
| SVM | 9456 | TF-IDF | 5188 | 2212 | 267 | 346 | 0.923 | 0.923 | 0.923 | 0.655 |
| RF | 244 | BOW | 2338 | 5270 | 223 | 182 | 0.950 | 0.949 | 0.950 | 0.787 |
| RF | 244 | TF-IDF | 2332 | 5264 | 223 | 194 | 0.948 | 0.948 | 0.948 | 0.778 |
| RF | 1226 | BOW | 2407 | 5223 | 208 | 175 | 0.952 | 0.952 | 0.952 | 0.797 |
| RF | 1226 | TF-IDF | 2361 | 5281 | 209 | 162 | **0.954** | **0.954** | **0.954** | 0.807 |
| RF | 4634 | BOW | 2378 | 5254 | 213 | 168 | 0.953 | 0.952 | 0.953 | 0.801 |
| RF | 4634 | TF-IDF | 2374 | 5232 | 196 | 211 | 0.949 | 0.949 | 0.949 | 0.772 |
| RF | 9456 | BOW | 2399 | 5228 | 188 | 198 | 0.952 | 0.952 | 0.952 | 0.784 |
| RF | 9456 | TF-IDF | 2419 | 5174 | 228 | 192 | 0.948 | 0.948 | 0.948 | 0.778 |

### 10.1.5.2 Using allergenic ingredients only

| Algo | Vocab | TextT | TP | TN | FP | FN | Pr | Re | F1 | Alpha |
|---|---|---|---|---|---|---|---|---|---|---|
| NN | 75 | BOW | 2258 | 2695 | 251 | 289 | 0.902 | 0.902 | 0.902 | 0.579 |
| NN | 75 | TF-IDF | 2254 | 2749 | 206 | 284 | 0.911 | 0.911 | 0.911 | 0.600 |
| NN | 298 | BOW | 2340 | 2674 | 218 | 261 | 0.913 | 0.913 | 0.913 | 0.618 |
| NN | 298 | TF-IDF | 2384 | 2675 | 190 | 244 | 0.921 | 0.921 | 0.921 | 0.645 |
| NN | 1225 | BOW | 2360 | 2711 | 210 | 212 | 0.923 | 0.923 | 0.923 | 0.671 |
| NN | 1225 | TF-IDF | 2375 | 2663 | 200 | 255 | 0.917 | 0.917 | 0.917 | 0.631 |
| NN | 3432 | BOW | 2332 | 2758 | 213 | 190 | **0.927** | **0.927** | **0.927** | **0.693** |
| NN | 3432 | TF-IDF | 2370 | 2699 | 188 | 236 | 0.923 | 0.923 | 0.923 | 0.654 |

## 10.1.6 Sulphur

### 10.1.6.1 Using list of ingredients

| Algo | Vocab | TextT | TP | TN | FP | FN | Pr | Re | F1 | Alpha |
|---|---|---|---|---|---|---|---|---|---|---|
| NN | 244 | BOW | 346 | 7476 | 60 | 130 | 0.975 | 0.976 | 0.975 | 0.873 |
| NN | 244 | TF-IDF | 386 | 7443 | 63 | 121 | 0.976 | 0.977 | 0.976 | 0.880 |
| NN | 1226 | BOW | 455 | 7463 | 22 | 73 | 0.988 | 0.988 | 0.988 | 0.931 |
| NN | 1226 | TF-IDF | 406 | 7502 | 23 | 82 | 0.987 | 0.987 | 0.987 | 0.923 |
| NN | 4634 | BOW | 438 | 7480 | 26 | 69 | 0.988 | 0.988 | 0.988 | 0.933 |
| NN | 4634 | TF-IDF | 443 | 7472 | 31 | 67 | 0.988 | 0.988 | 0.988 | **0.934** |
| NN | 9456 | BOW | 432 | 7467 | 20 | 93 | 0.985 | 0.986 | 0.985 | 0.915 |

| Algo | Vocab | TextT | TP | TN | FP | FN | Pr | Re | F1 | Alpha |
|---|---|---|---|---|---|---|---|---|---|---|
| NN | 9456 | TF-IDF | 439 | 7472 | 31 | 71 | 0.987 | 0.987 | 0.987 | 0.930 |
| SVM | 244 | BOW | 7514 | 261 | 2 | 236 | 0.971 | 0.970 | 0.966 | 0.805 |
| SVM | 244 | TF-IDF | 7492 | 272 | 1 | 247 | 0.970 | 0.969 | 0.964 | 0.797 |
| SVM | 1226 | BOW | 7442 | 424 | 77 | 69 | 0.982 | 0.982 | 0.982 | 0.919 |
| SVM | 1226 | TF-IDF | 7445 | 446 | 43 | 79 | 0.984 | 0.985 | 0.985 | 0.920 |
| SVM | 4634 | BOW | 7371 | 425 | 142 | 75 | 0.975 | 0.973 | 0.974 | 0.896 |
| SVM | 4634 | TF-IDF | 7427 | 420 | 94 | 72 | 0.980 | 0.979 | 0.979 | 0.912 |
| SVM | 9456 | BOW | 7443 | 396 | 96 | 78 | 0.979 | 0.978 | 0.978 | 0.906 |
| SVM | 9456 | TF-IDF | 7458 | 406 | 69 | 80 | 0.981 | 0.981 | 0.981 | 0.912 |
| RF | 244 | BOW | 346 | 7511 | 15 | 141 | 0.980 | 0.981 | 0.979 | 0.877 |
| RF | 244 | TF-IDF | 307 | 7512 | 6 | 188 | 0.976 | 0.976 | 0.973 | 0.842 |
| RF | 1226 | BOW | 430 | 7487 | 18 | 78 | 0.988 | 0.988 | 0.988 | 0.928 |
| RF | 1226 | TF-IDF | 395 | 7530 | 7 | 81 | **0.989** | **0.989** | **0.989** | 0.929 |
| RF | 4634 | BOW | 387 | 7525 | 8 | 93 | 0.987 | 0.987 | 0.987 | 0.918 |
| RF | 4634 | TF-IDF | 411 | 7510 | 6 | 86 | 0.988 | **0.989** | 0.988 | 0.925 |
| RF | 9456 | BOW | 418 | 7495 | 8 | 91 | 0.987 | 0.988 | 0.987 | 0.920 |
| RF | 9456 | TF-IDF | 420 | 7489 | 8 | 95 | 0.987 | 0.987 | 0.986 | 0.917 |

### 10.1.6.2 Using allergenic ingredients only

| Algo | Vocab | TextT | TP | TN | FP | FN | Pr | Re | F1 | Alpha |
|---|---|---|---|---|---|---|---|---|---|---|
| NN | 75 | BOW | 271 | 4998 | 25 | 199 | 0.958 | 0.959 | 0.955 | 0.755 |
| NN | 75 | TF-IDF | 277 | 4974 | 26 | 216 | 0.954 | 0.956 | 0.951 | 0.736 |
| NN | 298 | BOW | 414 | 4982 | 12 | 85 | 0.982 | 0.982 | 0.982 | 0.890 |
| NN | 298 | TF-IDF | 399 | 4970 | 24 | 99 | 0.977 | 0.977 | 0.977 | 0.869 |
| NN | 1225 | BOW | 414 | 4998 | 15 | 66 | **0.985** | **0.985** | **0.985** | 0.912 |
| NN | 1225 | TF-IDF | 416 | 4996 | 20 | 61 | **0.985** | **0.985** | **0.985** | **0.916** |
| NN | 3432 | BOW | 434 | 4967 | 19 | 72 | 0.983 | 0.983 | 0.983 | 0.903 |
| NN | 3432 | TF-IDF | 421 | 4965 | 25 | 82 | 0.980 | 0.981 | 0.980 | 0.888 |

## 10.1.7 Eggs

### *10.1.7.1 Using list of ingredients*

| Algo | Vocab | TextT | TP | TN | FP | FN | Pr | Re | F1 | Alpha |
|---|---|---|---|---|---|---|---|---|---|---|
| NN | 244 | BOW | 1711 | 5900 | 171 | 231 | 0.949 | 0.950 | 0.950 | 0.764 |
| NN | 244 | TF-IDF | 1728 | 5857 | 205 | 223 | 0.946 | 0.947 | 0.947 | 0.761 |
| NN | 1226 | BOW | 1775 | 5996 | 82 | 160 | 0.970 | 0.970 | 0.970 | 0.843 |
| NN | 1226 | TF-IDF | 1798 | 5965 | 124 | 126 | 0.969 | 0.969 | 0.969 | 0.859 |
| NN | 4634 | BOW | 1819 | 5957 | 76 | 161 | 0.970 | 0.970 | 0.970 | 0.844 |
| NN | 4634 | TF-IDF | 1795 | 5981 | 99 | 138 | 0.970 | 0.970 | 0.970 | 0.856 |
| NN | 9456 | BOW | 1856 | 5949 | 89 | 119 | **0.974** | **0.974** | **0.974** | **0.874** |
| NN | 9456 | TF-IDF | 1821 | 5963 | 105 | 124 | 0.971 | 0.971 | 0.971 | 0.866 |
| SVM | 244 | BOW | 5875 | 1529 | 173 | 436 | 0.923 | 0.924 | 0.922 | 0.617 |
| SVM | 244 | TF-IDF | 5866 | 1525 | 148 | 474 | 0.922 | 0.922 | 0.920 | 0.598 |
| SVM | 1226 | BOW | 5912 | 1804 | 110 | 187 | 0.963 | 0.963 | 0.963 | 0.814 |
| SVM | 1226 | TF-IDF | 5972 | 1731 | 96 | 214 | 0.961 | 0.961 | 0.961 | 0.797 |
| SVM | 4634 | BOW | 5869 | 1728 | 229 | 187 | 0.949 | 0.948 | 0.948 | 0.782 |
| SVM | 4634 | TF-IDF | 5887 | 1776 | 168 | 182 | 0.956 | 0.956 | 0.956 | 0.802 |
| SVM | 9456 | BOW | 5811 | 1742 | 233 | 227 | 0.943 | 0.943 | 0.943 | 0.750 |
| SVM | 9456 | TF-IDF | 5903 | 1741 | 156 | 213 | 0.954 | 0.954 | 0.954 | 0.782 |
| RF | 244 | BOW | 1745 | 5944 | 107 | 217 | 0.959 | 0.960 | 0.959 | 0.792 |
| RF | 244 | TF-IDF | 1719 | 5918 | 129 | 247 | 0.953 | 0.953 | 0.953 | 0.763 |
| RF | 1226 | BOW | 1784 | 5973 | 69 | 187 | 0.968 | 0.968 | 0.968 | 0.825 |
| RF | 1226 | TF-IDF | 1769 | 5975 | 100 | 169 | 0.966 | 0.966 | 0.966 | 0.831 |
| RF | 4634 | BOW | 1786 | 5997 | 78 | 152 | 0.971 | 0.971 | 0.971 | 0.851 |
| RF | 4634 | TF-IDF | 1756 | 5999 | 87 | 171 | 0.968 | 0.968 | 0.968 | 0.833 |
| RF | 9456 | BOW | 1806 | 5959 | 81 | 167 | 0.969 | 0.969 | 0.969 | 0.838 |
| RF | 9456 | TF-IDF | 1792 | 5947 | 95 | 179 | 0.966 | 0.966 | 0.966 | 0.825 |

### *10.1.7.2 Nur allergene Zutaten*

| Algo | Vocab | TextT | TP | TN | FP | FN | Pr | Re | F1 | Alpha |
|---|---|---|---|---|---|---|---|---|---|---|
| NN | 75 | BOW | 1619 | 3387 | 135 | 352 | 0.912 | 0.911 | 0.910 | 0.561 |
| NN | 75 | TF-IDF | 1657 | 3350 | 141 | 345 | 0.912 | 0.912 | 0.910 | 0.566 |
| NN | 298 | BOW | 1667 | 3476 | 111 | 239 | 0.936 | 0.936 | 0.936 | 0.680 |
| NN | 298 | TF-IDF | 1737 | 3399 | 85 | 272 | 0.936 | 0.935 | 0.934 | 0.656 |
| NN | 1225 | BOW | 1771 | 3414 | 85 | 223 | **0.944** | **0.944** | **0.943** | **0.707** |
| NN | 1225 | TF-IDF | 1682 | 3486 | 70 | 255 | 0.942 | 0.941 | 0.940 | 0.679 |
| NN | 3432 | BOW | 1777 | 3375 | 92 | 249 | 0.938 | 0.938 | 0.937 | 0.677 |
| NN | 3432 | TF-IDF | 1715 | 3424 | 101 | 253 | 0.936 | 0.936 | 0.935 | 0.669 |

### 10.1.8 Fish

#### *10.1.8.1 Using list of ingredients*

| Algo | Vocab | TextT | TP | TN | FP | FN | Pr | Re | F1 | Alpha |
|---|---|---|---|---|---|---|---|---|---|---|
| NN | 244 | BOW | 280 | 7375 | 46 | 312 | 0.952 | 0.955 | 0.949 | 0.737 |
| NN | 244 | TF-IDF | 282 | 7358 | 79 | 294 | 0.949 | 0.953 | 0.948 | 0.741 |
| NN | 1226 | BOW | 406 | 7427 | 26 | 154 | 0.977 | 0.978 | 0.976 | 0.863 |
| NN | 1226 | TF-IDF | 389 | 7445 | 48 | 131 | 0.977 | 0.978 | 0.977 | 0.876 |
| NN | 4634 | BOW | 476 | 7423 | 28 | 86 | 0.985 | 0.986 | 0.985 | 0.919 |
| NN | 4634 | TF-IDF | 443 | 7431 | 24 | 115 | 0.982 | 0.983 | 0.982 | 0.896 |
| NN | 9456 | BOW | 486 | 7417 | 23 | 87 | 0.986 | 0.986 | 0.986 | 0.919 |
| NN | 9456 | TF-IDF | 490 | 7422 | 27 | 74 | **0.987** | **0.987** | **0.987** | **0.929** |
| SVM | 244 | BOW | 7450 | 72 | 24 | 467 | 0.928 | 0.939 | 0.918 | 0.633 |
| SVM | 244 | TF-IDF | 7474 | 55 | 5 | 479 | 0.938 | 0.940 | 0.916 | 0.630 |
| SVM | 1226 | BOW | 7408 | 376 | 50 | 179 | 0.970 | 0.971 | 0.970 | 0.837 |
| SVM | 1226 | TF-IDF | 7435 | 383 | 38 | 157 | 0.975 | 0.976 | 0.974 | 0.858 |
| SVM | 4634 | BOW | 7314 | 480 | 111 | 108 | 0.973 | 0.973 | 0.973 | 0.877 |
| SVM | 4634 | TF-IDF | 7364 | 461 | 83 | 105 | 0.976 | 0.977 | 0.976 | 0.887 |
| SVM | 9456 | BOW | 7357 | 436 | 116 | 104 | 0.973 | 0.973 | 0.973 | 0.879 |
| SVM | 9456 | TF-IDF | 7373 | 463 | 77 | 100 | 0.978 | 0.978 | 0.978 | 0.893 |
| RF | 244 | BOW | 217 | 7440 | 30 | 326 | 0.953 | 0.956 | 0.948 | 0.731 |
| RF | 244 | TF-IDF | 184 | 7443 | 10 | 376 | 0.952 | 0.952 | 0.941 | 0.700 |

| Algo | Vocab | TextT | TP | TN | FP | FN | Pr | Re | F1 | Alpha |
|---|---|---|---|---|---|---|---|---|---|---|
| RF | 1226 | BOW | 386 | 7467 | 14 | 146 | 0.980 | 0.980 | 0.979 | 0.873 |
| RF | 1226 | TF-IDF | 394 | 7421 | 13 | 185 | 0.975 | 0.975 | 0.973 | 0.842 |
| RF | 4634 | BOW | 447 | 7441 | 13 | 112 | 0.984 | 0.984 | 0.984 | 0.901 |
| RF | 4634 | TF-IDF | 448 | 7445 | 10 | 110 | 0.985 | 0.985 | 0.984 | 0.904 |
| RF | 9456 | BOW | 430 | 7488 | 5 | 90 | 0.988 | 0.988 | 0.988 | 0.922 |
| RF | 9456 | TF-IDF | 432 | 7452 | 9 | 120 | 0.984 | 0.984 | 0.983 | 0.896 |

### *10.1.8.2 Using allergenic ingredients only*

| Algo | Vocab | TextT | TP | TN | FP | FN | Pr | Re | F1 | Alpha |
|---|---|---|---|---|---|---|---|---|---|---|
| NN | 75 | BOW | 189 | 4901 | 58 | 345 | 0.918 | 0.927 | 0.914 | 0.595 |
| NN | 75 | TF-IDF | 169 | 4902 | 56 | 366 | 0.913 | 0.923 | 0.909 | 0.576 |
| NN | 298 | BOW | 351 | 4925 | 43 | 174 | 0.959 | 0.960 | 0.958 | 0.776 |
| NN | 298 | TF-IDF | 382 | 4871 | 39 | 201 | 0.955 | 0.956 | 0.953 | 0.748 |
| NN | 1225 | BOW | 474 | 4879 | 11 | 129 | 0.975 | 0.975 | 0.973 | 0.839 |
| NN | 1225 | TF-IDF | 473 | 4910 | 13 | 97 | **0.980** | **0.980** | **0.979** | 0.875 |
| NN | 3432 | BOW | 436 | 4928 | 25 | 104 | 0.976 | 0.977 | 0.976 | 0.862 |
| NN | 3432 | TF-IDF | 469 | 4901 | 39 | 84 | 0.977 | 0.978 | 0.977 | **0.880** |

### 10.1.9 Nuts

### *10.1.9.1 Using list of ingredients*

| Algo | Vocab | TextT | TP | TN | FP | FN | Pr | Re | F1 | Alpha |
|---|---|---|---|---|---|---|---|---|---|---|
| NN | 244 | BOW | 1341 | 6372 | 113 | 187 | 0.962 | 0.963 | 0.962 | 0.813 |
| NN | 244 | TF-IDF | 1253 | 6400 | 132 | 228 | 0.954 | 0.955 | 0.954 | 0.777 |
| NN | 1226 | BOW | 1356 | 6386 | 108 | 163 | 0.966 | 0.966 | 0.966 | 0.834 |
| NN | 1226 | TF-IDF | 1287 | 6376 | 74 | 276 | 0.956 | 0.956 | 0.955 | 0.756 |
| NN | 4634 | BOW | 1354 | 6390 | 112 | 157 | 0.966 | 0.966 | 0.966 | 0.837 |
| NN | 4634 | TF-IDF | 1392 | 6353 | 117 | 151 | 0.966 | 0.967 | 0.966 | **0.841** |
| NN | 9456 | BOW | 1336 | 6397 | 96 | 184 | 0.965 | 0.965 | 0.965 | 0.820 |
| NN | 9456 | TF-IDF | 1348 | 6389 | 99 | 177 | 0.965 | 0.966 | 0.965 | 0.825 |
| SVM | 244 | BOW | 6384 | 1154 | 138 | 337 | 0.939 | 0.941 | 0.939 | 0.694 |
| SVM | 244 | TF-IDF | 6329 | 1194 | 110 | 380 | 0.938 | 0.939 | 0.937 | 0.671 |
| SVM | 1226 | BOW | 6387 | 1241 | 131 | 254 | 0.951 | 0.952 | 0.951 | 0.757 |
| SVM | 1226 | TF-IDF | 6353 | 1283 | 158 | 218 | 0.952 | 0.953 | 0.953 | 0.777 |

| Algo | Vocab | TextT | | | | | | | | |
|---|---|---|---|---|---|---|---|---|---|---|
| SVM | 4634 | BOW | 6273 | 1241 | 263 | 236 | 0.938 | 0.938 | 0.938 | 0.735 |
| SVM | 4634 | TF-IDF | 6244 | 1330 | 205 | 234 | 0.945 | 0.945 | 0.945 | 0.752 |
| SVM | 9456 | BOW | 6235 | 1223 | 301 | 254 | 0.932 | 0.931 | 0.931 | 0.712 |
| SVM | 9456 | TF-IDF | 6328 | 1273 | 172 | 240 | 0.948 | 0.949 | 0.948 | 0.757 |
| RF | 244 | BOW | 1285 | 6418 | 97 | 213 | 0.961 | 0.961 | 0.961 | 0.798 |
| RF | 244 | TF-IDF | 1363 | 6339 | 108 | 202 | 0.961 | 0.961 | 0.961 | 0.803 |
| RF | 1226 | BOW | 1309 | 6434 | 103 | 167 | 0.966 | 0.966 | 0.966 | 0.832 |
| RF | 1226 | TF-IDF | 1353 | 6356 | 102 | 201 | 0.961 | 0.962 | 0.962 | 0.805 |
| RF | 4634 | BOW | 1363 | 6395 | 95 | 160 | **0.968** | **0.968** | **0.968** | 0.840 |
| RF | 4634 | TF-IDF | 1324 | 6355 | 130 | 204 | 0.958 | 0.958 | 0.958 | 0.796 |
| RF | 9456 | BOW | 1346 | 6396 | 90 | 181 | 0.966 | 0.966 | 0.966 | 0.824 |
| RF | 9456 | TF-IDF | 1297 | 6412 | 116 | 187 | 0.961 | 0.962 | 0.962 | 0.813 |

### *10.1.9.2 Using allergenic ingredients only*

| Algo | Vocab | TextT | TP | TN | FP | FN | Pr | Re | F1 | Alpha |
|---|---|---|---|---|---|---|---|---|---|---|
| NN | 75 | BOW | 1139 | 3894 | 131 | 329 | 0.915 | 0.916 | 0.914 | 0.584 |
| NN | 75 | TF-IDF | 1157 | 3876 | 140 | 320 | 0.915 | 0.916 | 0.914 | 0.589 |
| NN | 298 | BOW | 1275 | 3787 | 153 | 277 | 0.920 | 0.922 | 0.920 | 0.626 |
| NN | 298 | TF-IDF | 1236 | 3831 | 127 | 299 | 0.922 | 0.922 | 0.921 | 0.614 |
| NN | 1225 | BOW | 1298 | 3831 | 154 | 210 | 0.933 | 0.934 | 0.933 | **0.694** |
| NN | 1225 | TF-IDF | 1307 | 3834 | 121 | 231 | **0.935** | **0.936** | **0.935** | 0.685 |
| NN | 3432 | BOW | 1269 | 3832 | 145 | 247 | 0.928 | 0.929 | 0.928 | 0.659 |
| NN | 3432 | TF-IDF | 1297 | 3807 | 148 | 240 | 0.928 | 0.929 | 0.928 | 0.665 |

## 10.1.10 Celery

### *10.1.10.1   Using list of ingredients*

| Algo | Vocab | TextT | TP | TN | FP | FN | Pr | Re | F1 | Alpha |
|---|---|---|---|---|---|---|---|---|---|---|
| NN | 244 | BOW | 842 | 6995 | 71 | 105 | 0.978 | 0.978 | 0.978 | 0.891 |
| NN | 244 | TF-IDF | 803 | 7019 | 53 | 138 | 0.976 | 0.976 | 0.976 | 0.869 |
| NN | 1226 | BOW | 809 | 7049 | 57 | 98 | 0.980 | 0.981 | 0.980 | 0.900 |
| NN | 1226 | TF-IDF | 831 | 6995 | 72 | 115 | 0.976 | 0.977 | 0.976 | 0.882 |
| NN | 4634 | BOW | 837 | 7034 | 50 | 92 | **0.982** | **0.982** | **0.982** | **0.907** |

| Algo | Vocab | TextT | TP | TN | FP | FN | Pr | Re | F1 | Alpha |
|---|---|---|---|---|---|---|---|---|---|---|
| NN | 4634 | TF-IDF | 794 | 7056 | 63 | 100 | 0.979 | 0.980 | 0.979 | 0.897 |
| NN | 9456 | BOW | 778 | 7087 | 35 | 113 | 0.981 | **0.982** | 0.981 | 0.894 |
| NN | 9456 | TF-IDF | 836 | 7023 | 64 | 90 | 0.981 | 0.981 | 0.981 | 0.905 |
| SVM | 244 | BOW | 7006 | 777 | 71 | 159 | 0.971 | 0.971 | 0.971 | 0.847 |
| SVM | 244 | TF-IDF | 6996 | 779 | 65 | 173 | 0.970 | 0.970 | 0.970 | 0.838 |
| SVM | 1226 | BOW | 6987 | 794 | 87 | 145 | 0.970 | 0.971 | 0.971 | 0.854 |
| SVM | 1226 | TF-IDF | 6981 | 817 | 71 | 144 | 0.973 | 0.973 | 0.973 | 0.859 |
| SVM | 4634 | BOW | 6921 | 718 | 216 | 158 | 0.955 | 0.953 | 0.954 | 0.808 |
| SVM | 4634 | TF-IDF | 6975 | 770 | 127 | 141 | 0.966 | 0.967 | 0.966 | 0.846 |
| SVM | 9456 | BOW | 6924 | 738 | 196 | 155 | 0.957 | 0.956 | 0.957 | 0.816 |
| SVM | 9456 | TF-IDF | 6972 | 749 | 124 | 168 | 0.963 | 0.964 | 0.963 | 0.825 |
| RF | 244 | BOW | 786 | 7066 | 38 | 123 | 0.980 | 0.980 | 0.979 | 0.885 |
| RF | 244 | TF-IDF | 767 | 7064 | 38 | 144 | 0.977 | 0.977 | 0.977 | 0.868 |
| RF | 1226 | BOW | 782 | 7081 | 34 | 116 | 0.981 | 0.981 | 0.981 | 0.892 |
| RF | 1226 | TF-IDF | 734 | 7086 | 37 | 156 | 0.976 | 0.976 | 0.975 | 0.859 |
| RF | 4634 | BOW | 803 | 7046 | 36 | 128 | 0.979 | 0.980 | 0.979 | 0.882 |
| RF | 4634 | TF-IDF | 775 | 7065 | 42 | 131 | 0.978 | 0.978 | 0.978 | 0.878 |
| RF | 9456 | BOW | 757 | 7077 | 37 | 142 | 0.977 | 0.978 | 0.977 | 0.870 |
| RF | 9456 | TF-IDF | 771 | 7077 | 32 | 133 | 0.979 | 0.979 | 0.979 | 0.879 |

### 10.1.10.2 *Using allergenic ingredients only*

| Algo | Vocab | TextT | TP | TN | FP | FN | Pr | Re | F1 | Alpha |
|---|---|---|---|---|---|---|---|---|---|---|
| NN | 75 | BOW | 715 | 4485 | 77 | 216 | 0.945 | 0.947 | 0.945 | 0.717 |
| NN | 75 | TF-IDF | 676 | 4503 | 105 | 209 | 0.941 | 0.943 | 0.941 | 0.714 |
| NN | 298 | BOW | 725 | 4495 | 52 | 221 | 0.950 | 0.950 | 0.948 | 0.721 |
| NN | 298 | TF-IDF | 706 | 4557 | 37 | 193 | 0.958 | 0.958 | 0.956 | 0.757 |
| NN | 1225 | BOW | 748 | 4530 | 46 | 169 | **0.960** | **0.961** | **0.960** | **0.780** |
| NN | 1225 | TF-IDF | 683 | 4577 | 47 | 186 | 0.957 | 0.958 | 0.956 | 0.761 |
| NN | 3432 | BOW | 762 | 4494 | 62 | 175 | 0.956 | 0.957 | 0.956 | 0.767 |
| NN | 3432 | TF-IDF | 722 | 4537 | 63 | 171 | 0.957 | 0.957 | 0.956 | 0.771 |

## 10.1.11 Mustard

### *10.1.11.1    Using list of ingredients*

| Algo | Vocab | TextT | TP | TN | FP | FN | Pr | Re | F1 | Alpha |
|------|-------|-------|------|------|-----|-----|-------|-------|-------|-------|
| NN | 244 | BOW | 987 | 6822 | 71 | 133 | 0.974 | 0.975 | 0.974 | 0.868 |
| NN | 244 | TF-IDF | 920 | 6874 | 57 | 162 | 0.972 | 0.973 | 0.972 | 0.848 |
| NN | 1226 | BOW | 911 | 6950 | 67 | 85 | 0.981 | 0.981 | 0.981 | **0.908** |
| NN | 1226 | TF-IDF | 967 | 6895 | 30 | 121 | 0.981 | 0.981 | 0.981 | 0.889 |
| NN | 4634 | BOW | 962 | 6913 | 49 | 89 | 0.983 | **0.983** | **0.983** | 0.910 |
| NN | 4634 | TF-IDF | 966 | 6889 | 45 | 113 | 0.980 | 0.980 | 0.980 | 0.891 |
| NN | 9456 | BOW | 961 | 6915 | 36 | 101 | 0.983 | **0.983** | **0.983** | 0.904 |
| NN | 9456 | TF-IDF | 931 | 6920 | 56 | 106 | 0.980 | 0.980 | 0.980 | 0.894 |
| SVM | 244 | BOW | 6923 | 796 | 60 | 234 | 0.962 | 0.963 | 0.962 | 0.791 |
| SVM | 244 | TF-IDF | 6894 | 795 | 76 | 248 | 0.958 | 0.960 | 0.958 | 0.776 |
| SVM | 1226 | BOW | 6846 | 927 | 74 | 166 | 0.969 | 0.970 | 0.969 | 0.841 |
| SVM | 1226 | TF-IDF | 6959 | 846 | 73 | 135 | 0.974 | 0.974 | 0.974 | 0.866 |
| SVM | 4634 | BOW | 6788 | 911 | 171 | 143 | 0.961 | 0.961 | 0.961 | 0.832 |
| SVM | 4634 | TF-IDF | 6850 | 897 | 112 | 154 | 0.966 | 0.967 | 0.967 | 0.840 |
| SVM | 9456 | BOW | 6756 | 911 | 188 | 158 | 0.957 | 0.957 | 0.957 | 0.816 |
| SVM | 9456 | TF-IDF | 6826 | 915 | 124 | 148 | 0.966 | 0.966 | 0.966 | 0.841 |
| RF | 244 | BOW | 881 | 6940 | 25 | 167 | 0.976 | 0.976 | 0.975 | 0.853 |
| RF | 244 | TF-IDF | 906 | 6922 | 28 | 157 | 0.977 | 0.977 | 0.976 | 0.860 |
| RF | 1226 | BOW | 941 | 6912 | 27 | 133 | 0.980 | 0.980 | 0.980 | 0.880 |
| RF | 1226 | TF-IDF | 908 | 6940 | 25 | 140 | 0.979 | 0.979 | 0.979 | 0.875 |
| RF | 4634 | BOW | 916 | 6961 | 15 | 121 | **0.983** | **0.983** | **0.983** | 0.893 |
| RF | 4634 | TF-IDF | 903 | 6973 | 25 | 112 | **0.983** | **0.983** | **0.983** | 0.898 |
| RF | 9456 | BOW | 936 | 6934 | 22 | 121 | 0.982 | 0.982 | 0.982 | 0.891 |
| RF | 9456 | TF-IDF | 929 | 6912 | 23 | 149 | 0.978 | 0.979 | 0.978 | 0.868 |

### *10.1.11.2    Using allergenic ingredients only*

| Algo | Vocab | TextT | TP | TN | FP | FN | Pr | Re | F1 | Alpha |
|------|-------|-------|------|------|-----|-----|-------|-------|-------|-------|
| NN | 75 | BOW | 842 | 4392 | 45 | 214 | 0.953 | 0.953 | 0.951 | 0.731 |
| NN | 75 | TF-IDF | 850 | 4408 | 49 | 186 | 0.957 | 0.957 | 0.956 | 0.760 |
| NN | 298 | BOW | 891 | 4371 | 59 | 172 | 0.957 | 0.958 | 0.957 | 0.771 |

| Algo | Vocab | TextT | TP | TN | FP | FN | Pr | Re | F1 | Alpha |
|---|---|---|---|---|---|---|---|---|---|---|
| NN | 298 | TF-IDF | 885 | 4388 | 63 | 157 | 0.959 | 0.960 | 0.959 | 0.787 |
| NN | 1225 | BOW | 879 | 4399 | 65 | 150 | 0.960 | 0.961 | 0.960 | 0.794 |
| NN | 1225 | TF-IDF | 903 | 4401 | 39 | 150 | **0.965** | **0.966** | **0.965** | **0.804** |
| NN | 3432 | BOW | 889 | 4412 | 39 | 153 | 0.965 | 0.965 | 0.964 | 0.800 |
| NN | 3432 | TF-IDF | 926 | 4355 | 53 | 159 | 0.961 | 0.961 | 0.961 | 0.788 |

## 10.1.12 Peanuts

### *10.1.12.1   Using list of ingredients*

| Algo | Vocab | TextT | TP | TN | FP | FN | Pr | Re | F1 | Alpha |
|---|---|---|---|---|---|---|---|---|---|---|
| NN | 244 | BOW | 416 | 7410 | 55 | 132 | 0.976 | 0.977 | 0.976 | 0.873 |
| NN | 244 | TF-IDF | 387 | 7423 | 46 | 157 | 0.973 | 0.975 | 0.973 | 0.856 |
| NN | 1226 | BOW | 439 | 7433 | 39 | 102 | 0.982 | 0.982 | 0.982 | 0.902 |
| NN | 1226 | TF-IDF | 491 | 7384 | 53 | 85 | 0.982 | 0.983 | 0.983 | 0.912 |
| NN | 4634 | BOW | 431 | 7434 | 45 | 103 | 0.981 | 0.982 | 0.981 | 0.900 |
| NN | 4634 | TF-IDF | 470 | 7394 | 47 | 102 | 0.981 | 0.981 | 0.981 | 0.900 |
| NN | 9456 | BOW | 456 | 7425 | 46 | 86 | **0.983** | **0.984** | **0.983** | **0.914** |
| NN | 9456 | TF-IDF | 435 | 7416 | 57 | 105 | 0.979 | 0.980 | 0.979 | 0.895 |
| SVM | 244 | BOW | 7465 | 262 | 25 | 261 | 0.963 | 0.964 | 0.959 | 0.780 |
| SVM | 244 | TF-IDF | 7453 | 270 | 17 | 273 | 0.963 | 0.964 | 0.959 | 0.773 |
| SVM | 1226 | BOW | 7423 | 391 | 58 | 141 | 0.974 | 0.975 | 0.974 | 0.865 |
| SVM | 1226 | TF-IDF | 7392 | 416 | 57 | 148 | 0.973 | 0.974 | 0.973 | 0.860 |
| SVM | 4634 | BOW | 7275 | 394 | 200 | 144 | 0.959 | 0.957 | 0.958 | 0.823 |
| SVM | 4634 | TF-IDF | 7331 | 418 | 115 | 149 | 0.966 | 0.967 | 0.967 | 0.843 |
| SVM | 9456 | BOW | 7214 | 445 | 223 | 131 | 0.959 | 0.956 | 0.957 | 0.827 |
| SVM | 9456 | TF-IDF | 7368 | 396 | 120 | 129 | 0.969 | 0.969 | 0.969 | 0.858 |
| RF | 244 | BOW | 394 | 7424 | 22 | 173 | 0.975 | 0.976 | 0.974 | 0.849 |
| RF | 244 | TF-IDF | 349 | 7452 | 16 | 196 | 0.973 | 0.974 | 0.971 | 0.833 |
| RF | 1226 | BOW | 442 | 7400 | 22 | 149 | 0.978 | 0.979 | 0.977 | 0.869 |
| RF | 1226 | TF-IDF | 408 | 7444 | 22 | 139 | 0.979 | 0.980 | 0.979 | 0.877 |
| RF | 4634 | BOW | 402 | 7468 | 16 | 127 | 0.982 | 0.982 | 0.981 | 0.888 |
| RF | 4634 | TF-IDF | 369 | 7484 | 14 | 146 | 0.980 | 0.980 | 0.979 | 0.873 |
| RF | 9456 | BOW | 409 | 7431 | 20 | 153 | 0.978 | 0.978 | 0.977 | 0.866 |
| RF | 9456 | TF-IDF | 362 | 7470 | 21 | 160 | 0.977 | 0.977 | 0.976 | 0.860 |

### 10.1.12.2 Using allergenic ingredients only

| Algo | Vocab | TextT | TP | TN | FP | FN | Pr | Re | F1 | Alpha |
|------|-------|-------|-----|------|-----|-----|-------|-------|-------|-------|
| NN | 75 | BOW | 370 | 4855 | 78 | 190 | 0.948 | 0.951 | 0.949 | 0.744 |
| NN | 75 | TF-IDF | 343 | 4896 | 78 | 176 | 0.951 | 0.954 | 0.952 | 0.760 |
| NN | 298 | BOW | 370 | 4900 | 66 | 157 | 0.957 | 0.959 | 0.958 | 0.785 |
| NN | 298 | TF-IDF | 349 | 4941 | 38 | 165 | **0.962** | **0.963** | **0.961** | 0.787 |
| NN | 1225 | BOW | 391 | 4884 | 57 | 161 | 0.959 | 0.960 | 0.958 | 0.784 |
| NN | 1225 | TF-IDF | 391 | 4889 | 69 | 144 | 0.960 | 0.961 | 0.960 | **0.799** |
| NN | 3432 | BOW | 386 | 4876 | 73 | 158 | 0.956 | 0.958 | 0.956 | 0.782 |
| NN | 3432 | TF-IDF | 413 | 4831 | 114 | 135 | 0.954 | 0.955 | 0.954 | 0.791 |

## 10.1.13 Sesame

### 10.1.13.1 Using list of ingredients

| Algo | Vocab | TextT | TP | TN | FP | FN | Pr | Re | F1 | Alpha |
|------|-------|-------|------|------|-----|-----|-------|-------|-------|-------|
| NN | 244 | BOW | 439 | 7380 | 59 | 135 | 0.975 | 0.976 | 0.975 | 0.870 |
| NN | 244 | TF-IDF | 414 | 7385 | 83 | 131 | 0.972 | 0.973 | 0.973 | 0.866 |
| NN | 1226 | BOW | 443 | 7424 | 46 | 100 | **0.981** | **0.982** | **0.981** | **0.902** |
| NN | 1226 | TF-IDF | 468 | 7373 | 67 | 105 | 0.978 | 0.979 | 0.978 | 0.892 |
| NN | 4634 | BOW | 455 | 7412 | 28 | 118 | 0.981 | 0.982 | 0.981 | 0.892 |
| NN | 4634 | TF-IDF | 486 | 7353 | 72 | 102 | 0.978 | 0.978 | 0.978 | 0.893 |
| NN | 9456 | BOW | 454 | 7400 | 44 | 115 | 0.980 | 0.980 | 0.980 | 0.890 |
| NN | 9456 | TF-IDF | 494 | 7350 | 55 | 114 | 0.978 | 0.979 | 0.978 | 0.888 |
| SVM | 244 | BOW | 7388 | 327 | 48 | 250 | 0.960 | 0.963 | 0.959 | 0.782 |
| SVM | 244 | TF-IDF | 7404 | 357 | 45 | 207 | 0.967 | 0.969 | 0.966 | 0.816 |
| SVM | 1226 | BOW | 7346 | 424 | 83 | 160 | 0.968 | 0.970 | 0.969 | 0.843 |
| SVM | 1226 | TF-IDF | 7375 | 449 | 75 | 114 | 0.976 | 0.976 | 0.976 | 0.882 |
| SVM | 4634 | BOW | 7298 | 424 | 157 | 134 | 0.964 | 0.964 | 0.964 | 0.843 |
| SVM | 4634 | TF-IDF | 7290 | 464 | 108 | 151 | 0.967 | 0.968 | 0.967 | 0.843 |
| SVM | 9456 | BOW | 7256 | 409 | 199 | 149 | 0.958 | 0.957 | 0.957 | 0.820 |
| SVM | 9456 | TF-IDF | 7321 | 428 | 106 | 158 | 0.966 | 0.967 | 0.966 | 0.838 |
| RF | 244 | BOW | 411 | 7409 | 32 | 161 | 0.975 | 0.976 | 0.974 | 0.856 |
| RF | 244 | TF-IDF | 377 | 7416 | 29 | 191 | 0.972 | 0.973 | 0.970 | 0.833 |
| RF | 1226 | BOW | 444 | 7424 | 27 | 118 | **0.981** | 0.982 | 0.981 | 0.892 |

| Algo | Vocab | TextT | TP | TN | FP | FN | Pr | Re | F1 | Alpha |
|---|---|---|---|---|---|---|---|---|---|---|
| RF | 1226 | TF-IDF | 437 | 7411 | 30 | 135 | 0.979 | 0.979 | 0.978 | 0.878 |
| RF | 4634 | BOW | 440 | 7414 | 27 | 132 | 0.980 | 0.980 | 0.979 | 0.881 |
| RF | 4634 | TF-IDF | 396 | 7459 | 21 | 137 | 0.980 | 0.980 | 0.979 | 0.879 |
| RF | 9456 | BOW | 406 | 7409 | 24 | 174 | 0.975 | 0.975 | 0.974 | 0.848 |
| RF | 9456 | TF-IDF | 419 | 7390 | 19 | 185 | 0.974 | 0.975 | 0.973 | 0.841 |

### *10.1.13.2    Using allergenic ingredients only*

| Algo | Vocab | TextT | TP | TN | FP | FN | Pr | Re | F1 | Alpha |
|---|---|---|---|---|---|---|---|---|---|---|
| NN | 75 | BOW | 355 | 4866 | 67 | 205 | 0.948 | 0.950 | 0.947 | 0.732 |
| NN | 75 | TF-IDF | 388 | 4783 | 95 | 227 | 0.938 | 0.941 | 0.938 | 0.699 |
| NN | 298 | BOW | 386 | 4887 | 47 | 173 | 0.958 | 0.960 | 0.958 | 0.775 |
| NN | 298 | TF-IDF | 351 | 4899 | 60 | 183 | 0.953 | 0.956 | 0.953 | 0.759 |
| NN | 1225 | BOW | 415 | 4876 | 43 | 159 | **0.962** | **0.963** | **0.961** | **0.792** |
| NN | 1225 | TF-IDF | 409 | 4854 | 69 | 161 | 0.956 | 0.958 | 0.956 | 0.780 |
| NN | 3432 | BOW | 406 | 4865 | 66 | 156 | 0.958 | 0.960 | 0.958 | 0.786 |
| NN | 3432 | TF-IDF | 388 | 4848 | 92 | 165 | 0.951 | 0.953 | 0.952 | 0.766 |

## 10.1.14 Lupine

### *10.1.14.1    Using list of ingredients*

| Algo | Vocab | TextT | TP | TN | FP | FN | Pr | Re | F1 | Alpha |
|---|---|---|---|---|---|---|---|---|---|---|
| NN | 244 | BOW | 74 | 7836 | 9 | 94 | 0.986 | 0.987 | 0.985 | 0.917 |
| NN | 244 | TF-IDF | 85 | 7823 | 36 | 69 | 0.986 | 0.987 | 0.986 | 0.931 |
| NN | 1226 | BOW | 104 | 7824 | 25 | 60 | 0.989 | 0.989 | 0.989 | 0.941 |
| NN | 1226 | TF-IDF | 103 | 7817 | 42 | 51 | 0.988 | 0.988 | 0.988 | 0.944 |
| NN | 4634 | BOW | 102 | 7833 | 22 | 56 | 0.990 | 0.990 | 0.990 | 0.946 |
| NN | 4634 | TF-IDF | 104 | 7837 | 12 | 60 | 0.990 | **0.991** | 0.990 | 0.945 |
| NN | 9456 | BOW | 117 | 7824 | 27 | 45 | **0.991** | **0.991** | **0.991** | **0.953** |
| NN | 9456 | TF-IDF | 97 | 7839 | 23 | 54 | 0.990 | 0.990 | 0.990 | 0.947 |
| SVM | 244 | BOW | 7844 | 0 | 0 | 169 | 0.958 | 0.979 | 0.968 | 0.859 |
| SVM | 244 | TF-IDF | 7821 | 0 | 0 | 192 | 0.953 | 0.976 | 0.964 | 0.840 |
| SVM | 1226 | BOW | 7749 | 87 | 115 | 62 | 0.982 | 0.978 | 0.980 | 0.914 |
| SVM | 1226 | TF-IDF | 7824 | 77 | 36 | 76 | 0.984 | 0.986 | 0.985 | 0.925 |
| SVM | 4634 | BOW | 7729 | 90 | 120 | 74 | 0.979 | 0.976 | 0.977 | 0.903 |

| Algo | Vocab | TextT | TP | TN | FP | FN | Pr | Re | F1 | Alpha |
|---|---|---|---|---|---|---|---|---|---|---|
| SVM | 4634 | TF-IDF | 7782 | 96 | 69 | 66 | 0.983 | 0.983 | 0.983 | 0.924 |
| SVM | 9456 | BOW | 7754 | 85 | 112 | 62 | 0.982 | 0.978 | 0.980 | 0.915 |
| SVM | 9456 | TF-IDF | 7798 | 103 | 43 | 69 | 0.985 | 0.986 | 0.985 | 0.929 |
| RF | 244 | BOW | 62 | 7848 | 5 | 98 | 0.986 | 0.987 | 0.985 | 0.915 |
| RF | 244 | TF-IDF | 40 | 7864 | 5 | 104 | 0.985 | 0.986 | 0.983 | 0.910 |
| RF | 1226 | BOW | 77 | 7857 | 5 | 74 | 0.990 | 0.990 | 0.989 | 0.935 |
| RF | 1226 | TF-IDF | 84 | 7852 | 8 | 69 | 0.990 | 0.990 | 0.989 | 0.938 |
| RF | 4634 | BOW | 85 | 7851 | 6 | 71 | 0.990 | 0.990 | 0.989 | 0.937 |
| RF | 4634 | TF-IDF | 88 | 7855 | 3 | 67 | **0.991** | **0.991** | 0.990 | 0.942 |
| RF | 9456 | BOW | 77 | 7851 | 5 | 80 | 0.989 | 0.989 | 0.988 | 0.930 |
| RF | 9456 | TF-IDF | 63 | 7855 | 4 | 91 | 0.988 | 0.988 | 0.986 | 0.921 |

### *10.1.14.2    Using allergenic ingredients only*

| Algo | Vocab | TextT | TP | TN | FP | FN | Pr | Re | F1 | Alpha |
|---|---|---|---|---|---|---|---|---|---|---|
| NN | 75 | BOW | 52 | 5298 | 30 | 113 | 0.969 | 0.974 | 0.970 | 0.850 |
| NN | 75 | TF-IDF | 50 | 5325 | 13 | 105 | 0.975 | 0.979 | 0.974 | 0.866 |
| NN | 298 | BOW | 102 | 5290 | 27 | 74 | 0.980 | 0.982 | 0.980 | 0.897 |
| NN | 298 | TF-IDF | 70 | 5325 | 16 | 82 | 0.980 | 0.982 | 0.980 | 0.892 |
| NN | 1225 | BOW | 86 | 5320 | 16 | 71 | 0.983 | 0.984 | 0.983 | 0.905 |
| NN | 1225 | TF-IDF | 87 | 5327 | 18 | 61 | **0.984** | **0.986** | **0.984** | **0.917** |
| NN | 3432 | BOW | 88 | 5312 | 25 | 68 | 0.981 | 0.983 | 0.982 | 0.905 |
| NN | 3432 | TF-IDF | 108 | 5275 | 52 | 58 | 0.980 | 0.980 | 0.980 | 0.906 |

## 10.1.15 Molluscs

### *10.1.15.1    Using list of ingredients*

| Algo | Vocab | TextT | TP | TN | FP | FN | Pr | Re | F1 | Alpha |
|---|---|---|---|---|---|---|---|---|---|---|
| NN | 244 | BOW | 58 | 7858 | 23 | 74 | 0.986 | 0.988 | 0.986 | 0.930 |
| NN | 244 | TF-IDF | 43 | 7878 | 12 | 80 | 0.987 | 0.989 | 0.986 | 0.928 |
| NN | 1226 | BOW | 71 | 7858 | 28 | 56 | 0.989 | 0.990 | 0.989 | 0.944 |
| NN | 1226 | TF-IDF | 79 | 7855 | 23 | 56 | 0.989 | 0.990 | 0.989 | 0.945 |
| NN | 4634 | BOW | 92 | 7855 | 21 | 45 | 0.991 | 0.992 | 0.991 | 0.955 |
| NN | 4634 | TF-IDF | 83 | 7868 | 8 | 54 | 0.992 | 0.992 | 0.991 | 0.951 |
| NN | 9456 | BOW | 93 | 7862 | 21 | 37 | 0.992 | 0.993 | 0.993 | 0.962 |

| Algo | Vocab | TextT | TP | TN | FP | FN | Pr | Re | F1 | Alpha |
|---|---|---|---|---|---|---|---|---|---|---|
| NN | 9456 | TF-IDF | 108 | 7861 | 12 | 32 | **0.994** | **0.995** | **0.994** | **0.969** |
| SVM | 244 | BOW | 7901 | 0 | 0 | 112 | 0.972 | 0.986 | 0.979 | 0.905 |
| SVM | 244 | TF-IDF | 7887 | 0 | 0 | 126 | 0.969 | 0.984 | 0.976 | 0.893 |
| SVM | 1226 | BOW | 7854 | 51 | 30 | 78 | 0.984 | 0.987 | 0.985 | 0.925 |
| SVM | 1226 | TF-IDF | 7876 | 48 | 19 | 70 | 0.987 | 0.989 | 0.987 | 0.935 |
| SVM | 4634 | BOW | 7794 | 77 | 85 | 57 | 0.984 | 0.982 | 0.983 | 0.927 |
| SVM | 4634 | TF-IDF | 7849 | 66 | 46 | 52 | 0.987 | 0.988 | 0.988 | 0.942 |
| SVM | 9456 | BOW | 7826 | 72 | 58 | 57 | 0.986 | 0.986 | 0.986 | 0.934 |
| SVM | 9456 | TF-IDF | 7870 | 63 | 37 | 43 | 0.990 | 0.990 | 0.990 | 0.952 |
| RF | 244 | BOW | 19 | 7897 | 3 | 94 | 0.986 | 0.988 | 0.984 | 0.919 |
| RF | 244 | TF-IDF | 7 | 7876 | 5 | 125 | 0.978 | 0.984 | 0.977 | 0.893 |
| RF | 1226 | BOW | 49 | 7866 | 13 | 85 | 0.986 | 0.988 | 0.986 | 0.924 |
| RF | 1226 | TF-IDF | 58 | 7874 | 3 | 78 | 0.990 | 0.990 | 0.988 | 0.932 |
| RF | 4634 | BOW | 57 | 7893 | 3 | 60 | 0.992 | 0.992 | 0.991 | 0.948 |
| RF | 4634 | TF-IDF | 59 | 7895 | 6 | 53 | 0.992 | 0.993 | 0.992 | 0.953 |
| RF | 9456 | BOW | 69 | 7875 | 9 | 60 | 0.991 | 0.991 | 0.990 | 0.946 |
| RF | 9456 | TF-IDF | 59 | 7898 | 1 | 55 | 0.993 | 0.993 | 0.992 | 0.952 |

### *10.1.15.2  Using allergenic ingredients only*

| Algo | Vocab | TextT | TP | TN | FP | FN | Pr | Re | F1 | Alpha |
|---|---|---|---|---|---|---|---|---|---|---|
| NN | 75 | BOW | 30 | 5361 | 11 | 91 | 0.978 | 0.981 | 0.977 | 0.883 |
| NN | 75 | TF-IDF | 20 | 5358 | 24 | 91 | 0.973 | 0.979 | 0.975 | 0.878 |
| NN | 298 | BOW | 57 | 5360 | 16 | 60 | 0.984 | 0.986 | 0.985 | 0.919 |
| NN | 298 | TF-IDF | 45 | 5365 | 8 | 75 | 0.983 | 0.985 | 0.982 | 0.904 |
| NN | 1225 | BOW | 62 | 5359 | 23 | 49 | 0.986 | **0.987** | 0.986 | 0.929 |
| NN | 1225 | TF-IDF | 63 | 5341 | 22 | 67 | 0.982 | 0.984 | 0.982 | 0.908 |
| NN | 3432 | BOW | 67 | 5354 | 9 | 63 | 0.986 | **0.987** | 0.985 | 0.918 |
| NN | 3432 | TF-IDF | 65 | 5357 | 29 | 42 | **0.986** | **0.987** | **0.987** | **0.935** |

## 10.1.16 Crustaceans

### *10.1.16.1  Using list of ingredients*

| Algo | Vocab | TextT | TP | TN | FP | FN | Pr | Re | F1 | Alpha |
|---|---|---|---|---|---|---|---|---|---|---|
| NN | 244 | BOW | 95 | 7771 | 33 | 114 | 0.979 | 0.982 | 0.979 | 0.894 |

| Algo | Vocab | TextT | TP | TN | FP | FN | Pr | Re | F1 | Alpha |
|---|---|---|---|---|---|---|---|---|---|---|
| NN | 244 | TF-IDF | 54 | 7808 | 16 | 135 | 0.978 | 0.981 | 0.977 | 0.882 |
| NN | 1226 | BOW | 129 | 7819 | 19 | 46 | 0.991 | 0.992 | 0.992 | 0.955 |
| NN | 1226 | TF-IDF | 136 | 7799 | 20 | 58 | 0.990 | 0.990 | 0.990 | 0.944 |
| NN | 4634 | BOW | 140 | 7811 | 10 | 52 | 0.992 | 0.992 | 0.992 | 0.952 |
| NN | 4634 | TF-IDF | 130 | 7799 | 25 | 59 | 0.989 | 0.990 | 0.989 | 0.942 |
| NN | 9456 | BOW | 115 | 7844 | 13 | 41 | **0.993** | **0.993** | **0.993** | **0.961** |
| NN | 9456 | TF-IDF | 158 | 7783 | 17 | 55 | 0.991 | 0.991 | 0.991 | 0.948 |
| SVM | 244 | BOW | 7800 | 0 | 0 | 213 | 0.948 | 0.973 | 0.960 | 0.824 |
| SVM | 244 | TF-IDF | 7815 | 0 | 0 | 198 | 0.951 | 0.975 | 0.963 | 0.836 |
| SVM | 1226 | BOW | 7757 | 134 | 58 | 64 | 0.985 | 0.985 | 0.985 | 0.929 |
| SVM | 1226 | TF-IDF | 7793 | 117 | 21 | 82 | 0.986 | 0.987 | 0.986 | 0.924 |
| SVM | 4634 | BOW | 7709 | 123 | 110 | 71 | 0.980 | 0.977 | 0.978 | 0.908 |
| SVM | 4634 | TF-IDF | 7781 | 132 | 40 | 60 | 0.987 | 0.988 | 0.987 | 0.937 |
| SVM | 9456 | BOW | 7717 | 138 | 97 | 61 | 0.982 | 0.980 | 0.981 | 0.920 |
| SVM | 9456 | TF-IDF | 7773 | 127 | 50 | 63 | 0.985 | 0.986 | 0.986 | 0.932 |
| RF | 244 | BOW | 34 | 7820 | 6 | 153 | 0.978 | 0.980 | 0.974 | 0.870 |
| RF | 244 | TF-IDF | 15 | 7835 | 7 | 156 | 0.974 | 0.980 | 0.972 | 0.867 |
| RF | 1226 | BOW | 118 | 7824 | 10 | 61 | 0.991 | 0.991 | 0.990 | 0.945 |
| RF | 1226 | TF-IDF | 119 | 7803 | 11 | 80 | 0.988 | 0.989 | 0.987 | 0.928 |
| RF | 4634 | BOW | 118 | 7818 | 11 | 66 | 0.990 | 0.990 | 0.990 | 0.940 |
| RF | 4634 | TF-IDF | 118 | 7827 | 4 | 64 | 0.991 | 0.992 | 0.991 | 0.944 |
| RF | 9456 | BOW | 131 | 7799 | 9 | 74 | 0.989 | 0.990 | 0.989 | 0.934 |
| RF | 9456 | TF-IDF | 109 | 7825 | 6 | 73 | 0.990 | 0.990 | 0.989 | 0.936 |

### 10.1.16.2  Using allergenic ingredients only

| Algo | Vocab | TextT | TP | TN | FP | FN | Pr | Re | F1 | Alpha |
|---|---|---|---|---|---|---|---|---|---|---|
| NN | 75 | BOW | 41 | 5275 | 28 | 149 | 0.959 | 0.968 | 0.960 | 0.809 |
| NN | 75 | TF-IDF | 45 | 5272 | 29 | 147 | 0.960 | 0.968 | 0.961 | 0.811 |
| NN | 298 | BOW | 101 | 5280 | 19 | 93 | 0.978 | 0.980 | 0.977 | 0.878 |
| NN | 298 | TF-IDF | 76 | 5285 | 23 | 109 | 0.973 | 0.976 | 0.972 | 0.857 |
| NN | 1225 | BOW | 138 | 5249 | 15 | 91 | 0.980 | 0.981 | 0.979 | 0.882 |
| NN | 1225 | TF-IDF | 105 | 5304 | 17 | 67 | **0.984** | **0.985** | 0.983 | 0.910 |
| NN | 3432 | BOW | 117 | 5289 | 25 | 62 | 0.983 | 0.984 | **0.983** | **0.913** |
| NN | 3432 | TF-IDF | 130 | 5269 | 24 | 70 | 0.982 | 0.983 | 0.982 | 0.903 |

## 10.2 Results Classifer Chains

Results obtained using neural nets and averaging over ten runs with randomized chain order.

| Allergen | Voc | TT | $Pr_{macro}$ | $Re_{macro}$ | $F1_{macro}$ | $Pr_{micro}$ | $Re_{micro}$ | $F1_{micro}$ | Alpha |
|---|---|---|---|---|---|---|---|---|---|
| Milk | 1226 | BOW | 0.968 | 0.970 | 0.969 | 0.969 | 0.969 | 0.969 | 0.880 |
| Soybeans | 1226 | BOW | 0.964 | 0.958 | 0.961 | 0.971 | 0.971 | 0.971 | 0.859 |
| Gluten | 1226 | BOW | 0.942 | 0.949 | 0.946 | 0.953 | 0.953 | 0.953 | 0.810 |
| Sulphur | 1226 | BOW | 0.963 | 0.932 | 0.947 | 0.988 | 0.988 | 0.988 | 0.934 |
| Eggs | 1226 | BOW | 0.967 | 0.963 | 0.965 | 0.974 | 0.974 | 0.974 | 0.874 |
| Fish | 1226 | BOW | 0.969 | 0.933 | 0.950 | 0.987 | 0.987 | 0.987 | 0.929 |
| Nuts | 1226 | BOW | 0.950 | 0.942 | 0.946 | 0.967 | 0.967 | 0.967 | 0.841 |
| Celery | 1226 | BOW | 0.965 | 0.947 | 0.956 | 0.982 | 0.982 | 0.982 | 0.907 |
| Mustard | 1226 | BOW | 0.960 | 0.953 | 0.956 | 0.981 | 0.981 | 0.981 | 0.908 |
| Peanuts | 1226 | BOW | 0.948 | 0.918 | 0.932 | 0.984 | 0.984 | 0.984 | 0.914 |
| Sesame | 1226 | BOW | 0.946 | 0.905 | 0.924 | 0.982 | 0.982 | 0.982 | 0.902 |
| Lupine | 1226 | BOW | 0.903 | 0.859 | 0.880 | 0.991 | 0.991 | 0.991 | 0.953 |
| Molluscs | 1226 | BOW | 0.948 | 0.885 | 0.914 | 0.995 | 0.995 | 0.995 | 0.969 |
| Crustaceans | 1226 | BOW | 0.947 | 0.868 | 0.903 | 0.993 | 0.993 | 0.993 | 0.961 |
| Milk | 4634 | BOW | 0.968 | 0.970 | 0.969 | 0.969 | 0.969 | 0.969 | 0.880 |
| Soybeans | 4634 | BOW | 0.964 | 0.958 | 0.961 | 0.971 | 0.971 | 0.971 | 0.859 |
| Gluten | 4634 | BOW | 0.942 | 0.949 | 0.946 | 0.953 | 0.953 | 0.953 | 0.810 |
| Sulphur | 4634 | BOW | 0.963 | 0.932 | 0.947 | 0.988 | 0.988 | 0.988 | 0.934 |
| Eggs | 4634 | BOW | 0.967 | 0.963 | 0.965 | 0.974 | 0.974 | 0.974 | 0.874 |
| Fish | 4634 | BOW | 0.969 | 0.933 | 0.950 | 0.987 | 0.987 | 0.987 | 0.929 |
| Nuts | 4634 | BOW | 0.950 | 0.942 | 0.946 | 0.967 | 0.967 | 0.967 | 0.841 |
| Celery | 4634 | BOW | 0.965 | 0.947 | 0.956 | 0.982 | 0.982 | 0.982 | 0.907 |
| Mustard | 4634 | BOW | 0.960 | 0.953 | 0.956 | 0.981 | 0.981 | 0.981 | 0.908 |
| Peanuts | 4634 | BOW | 0.948 | 0.918 | 0.932 | 0.984 | 0.984 | 0.984 | 0.914 |
| Sesame | 4634 | BOW | 0.946 | 0.905 | 0.924 | 0.982 | 0.982 | 0.982 | 0.902 |
| Lupine | 4634 | BOW | 0.903 | 0.859 | 0.880 | 0.991 | 0.991 | 0.991 | 0.953 |
| Molluscs | 4634 | BOW | 0.948 | 0.885 | 0.914 | 0.995 | 0.995 | 0.995 | 0.969 |
| Crustaceans | 4634 | BOW | 0.947 | 0.868 | 0.903 | 0.993 | 0.993 | 0.993 | 0.961 |

| Allergen | Voc | TT | $Pr_{macro}$ | $Re_{macro}$ | $F1_{macro}$ | $Pr_{micro}$ | $Re_{micro}$ | $F1_{micro}$ | Alpha |
|---|---|---|---|---|---|---|---|---|---|
| Milk | 1226 | TF-IDF | 0.968 | 0.970 | 0.969 | 0.969 | 0.969 | 0.969 | 0.880 |
| Soybeans | 1226 | TF-IDF | 0.964 | 0.958 | 0.961 | 0.971 | 0.971 | 0.971 | 0.859 |
| Gluten | 1226 | TF-IDF | 0.942 | 0.949 | 0.946 | 0.953 | 0.953 | 0.953 | 0.810 |
| Sulphur | 1226 | TF-IDF | 0.963 | 0.932 | 0.947 | 0.988 | 0.988 | 0.988 | 0.934 |
| Eggs | 1226 | TF-IDF | 0.967 | 0.963 | 0.965 | 0.974 | 0.974 | 0.974 | 0.874 |
| Fish | 1226 | TF-IDF | 0.969 | 0.933 | 0.950 | 0.987 | 0.987 | 0.987 | 0.929 |
| Nuts | 1226 | TF-IDF | 0.950 | 0.942 | 0.946 | 0.967 | 0.967 | 0.967 | 0.841 |
| Celery | 1226 | TF-IDF | 0.965 | 0.947 | 0.956 | 0.982 | 0.982 | 0.982 | 0.907 |
| Mustard | 1226 | TF-IDF | 0.960 | 0.953 | 0.956 | 0.981 | 0.981 | 0.981 | 0.908 |
| Peanuts | 1226 | TF-IDF | 0.948 | 0.918 | 0.932 | 0.984 | 0.984 | 0.984 | 0.914 |
| Sesame | 1226 | TF-IDF | 0.946 | 0.905 | 0.924 | 0.982 | 0.982 | 0.982 | 0.902 |
| Lupine | 1226 | TF-IDF | 0.903 | 0.859 | 0.880 | 0.991 | 0.991 | 0.991 | 0.953 |
| Molluscs | 1226 | TF-IDF | 0.948 | 0.885 | 0.914 | 0.995 | 0.995 | 0.995 | 0.969 |
| Crustaceans | 1226 | TF-IDF | 0.947 | 0.868 | 0.903 | 0.993 | 0.993 | 0.993 | 0.961 |
| Milk | 4634 | TF-IDF | 0.968 | 0.970 | 0.969 | 0.969 | 0.969 | 0.969 | 0.880 |
| Soybeans | 4634 | TF-IDF | 0.964 | 0.958 | 0.961 | 0.971 | 0.971 | 0.971 | 0.859 |
| Gluten | 4634 | TF-IDF | 0.942 | 0.949 | 0.946 | 0.953 | 0.953 | 0.953 | 0.810 |
| Sulphur | 4634 | TF-IDF | 0.963 | 0.932 | 0.947 | 0.988 | 0.988 | 0.988 | 0.934 |
| Eggs | 4634 | TF-IDF | 0.967 | 0.963 | 0.965 | 0.974 | 0.974 | 0.974 | 0.874 |
| Fish | 4634 | TF-IDF | 0.969 | 0.933 | 0.950 | 0.987 | 0.987 | 0.987 | 0.929 |
| Nuts | 4634 | TF-IDF | 0.950 | 0.942 | 0.946 | 0.967 | 0.967 | 0.967 | 0.841 |
| Celery | 4634 | TF-IDF | 0.965 | 0.947 | 0.956 | 0.982 | 0.982 | 0.982 | 0.907 |
| Mustard | 4634 | TF-IDF | 0.960 | 0.953 | 0.956 | 0.981 | 0.981 | 0.981 | 0.908 |
| Peanuts | 4634 | TF-IDF | 0.948 | 0.918 | 0.932 | 0.984 | 0.984 | 0.984 | 0.914 |
| Sesame | 4634 | TF-IDF | 0.946 | 0.905 | 0.924 | 0.982 | 0.982 | 0.982 | 0.902 |
| Lupine | 4634 | TF-IDF | 0.903 | 0.859 | 0.880 | 0.991 | 0.991 | 0.991 | 0.953 |
| Molluscs | 4634 | TF-IDF | 0.948 | 0.885 | 0.914 | 0.995 | 0.995 | 0.995 | 0.969 |
| Crustaceans | 4634 | TF-IDF | 0.947 | 0.868 | 0.903 | 0.993 | 0.993 | 0.993 | 0.961 |